\documentclass[10pt,twocolumn,letterpaper]{article}
\usepackage{iccv}

\usepackage{authblk}
\usepackage{newtxtext}
\usepackage{newtxmath}
\usepackage{amsmath}
\usepackage{multirow}
\usepackage{graphicx}
\usepackage{xcolor}
\usepackage{pifont}
\usepackage{wrapfig}
\usepackage{subfig}
\usepackage{enumitem}
\usepackage{multibib}
\usepackage[export]{adjustbox}

\usepackage[pagebackref=true,breaklinks=true,colorlinks,bookmarks=false]{hyperref}
\usepackage{cleveref}
\setitemize{noitemsep,topsep=0pt,parsep=0pt,partopsep=0pt}
\setenumerate{noitemsep,topsep=0pt,parsep=0pt,partopsep=0pt}

\makeatletter
\renewcommand{\paragraph}{%
  \@startsection{paragraph}{4}%
  {\z@}{1ex \@plus .25ex \@minus .2ex}{-1em}%
  {\normalfont\normalsize\bfseries}%
}
\makeatother

\newcommand{\ignore}[1]{}

\newcommand{\OURS}{HoliCity}

\newcommand{\Xwgs}{\mathbf{X}_\mathrm{WGS}}
\newcommand{\Xcad}{\mathbf{X}_\mathrm{CAD}}
\newcommand{\X}{\mathbf{X}}
\newcommand{\x}{\mathbf{x}}

\iccvfinalcopy
\def\iccvPaperID{9795} %

\ificcvfinal\fi

\newcites{supp}{References of Supplementary Material}

\begin{document}

\setlength{\belowcaptionskip}{0pt}
\setlength{\abovecaptionskip}{1.2ex}
\addtolength{\floatsep}{-1ex}
\addtolength{\textfloatsep}{-2ex}
\addtolength{\dbltextfloatsep}{-1.5ex}
\setlength{\abovedisplayskip}{6pt}
\setlength{\belowdisplayskip}{6pt}
\newcommand{\simplify}[1]{}

\title{\OURS{}: A City-Scale Data Platform for Learning Holistic 3D Structures}

\author[1]{Yichao Zhou\thanks{This work is sponsored by a generous grant from Sony Research US.}}
\author[2]{Jingwei Huang}
\author[1]{Xili Dai}
\author[3]{Shichen Liu}
\author[4]{Linjie Luo}
\author[4]{Zhili Chen}
\author[1]{Yi Ma}
\affil[1]{University of California, Berkeley, California, USA}
\affil[2]{Stanford University, Stanford, California, USA}
\affil[3]{University of Southern California, Los Angeles, California, USA}
\affil[4]{Bytedance Research, Palo Alto, California, USA}

\maketitle

\begin{abstract}
We present HoliCity, a city-scale 3D dataset with rich structural information. %
Currently, this dataset has 6,300 real-world panoramas of resolution $13312 \times 6656$ that are accurately aligned with the CAD model of downtown London with an area of more than 20 km$^2$, in which the median reprojection error of the alignment of an average image is less than half a degree.
This dataset aims to be an all-in-one data platform for research of learning abstracted high-level holistic 3D structures that can be derived from city CAD models, e.g., corners, lines, wireframes, planes, and cuboids, with the ultimate goal of supporting real-world applications including city-scale reconstruction, localization, mapping and augmented reality.
The accurate alignment of the 3D CAD models and panoramas also benefits low-level 3D vision tasks such as surface normal estimation, as the surface normal extracted from previous LiDAR-based datasets is often noisy.
We conduct experiments to demonstrate the applications of HoliCity, such as predicting surface segmentation, normal maps, depth maps, and vanishing points, as well as test the generalizability of methods trained on HoliCity and other related datasets.
\end{abstract}

\begin{figure*}[t]
\vspace{-0.15in}
  \centering
  \setlength{\lineskip}{0pt}
  \subfloat[Bird's-eye view of the HoliCity CAD model
  \label{fig:teaser:CAD}]{%
    \includegraphics[height=.197\linewidth,trim=0 0 555 0,clip]{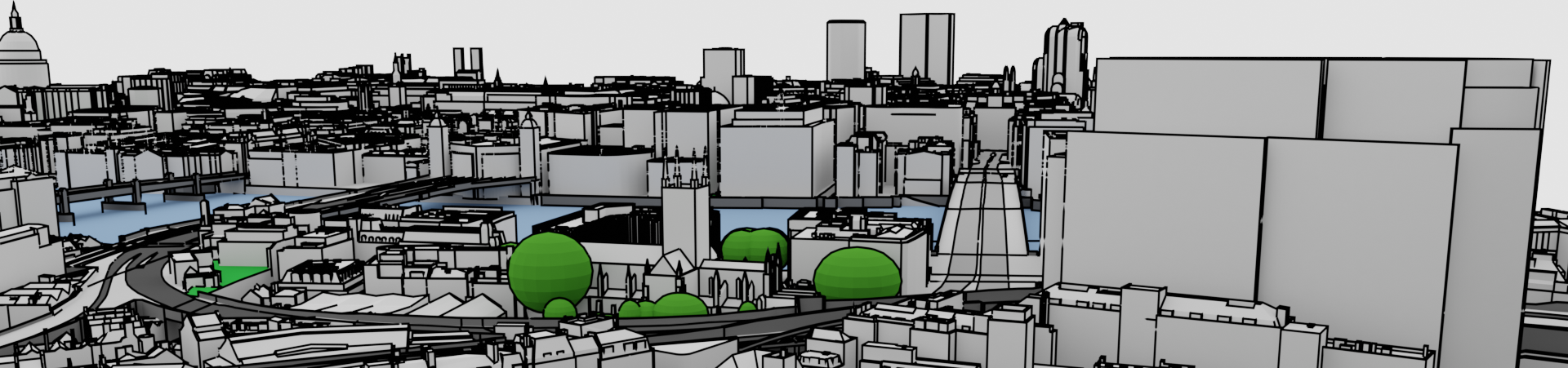}%
  }%
  \hfill%
  \subfloat[Viewpoint coverage
  \label{fig:teaser:satellite}]{%
    \includegraphics[height=.197\linewidth,trim=0 350 0 100,clip]{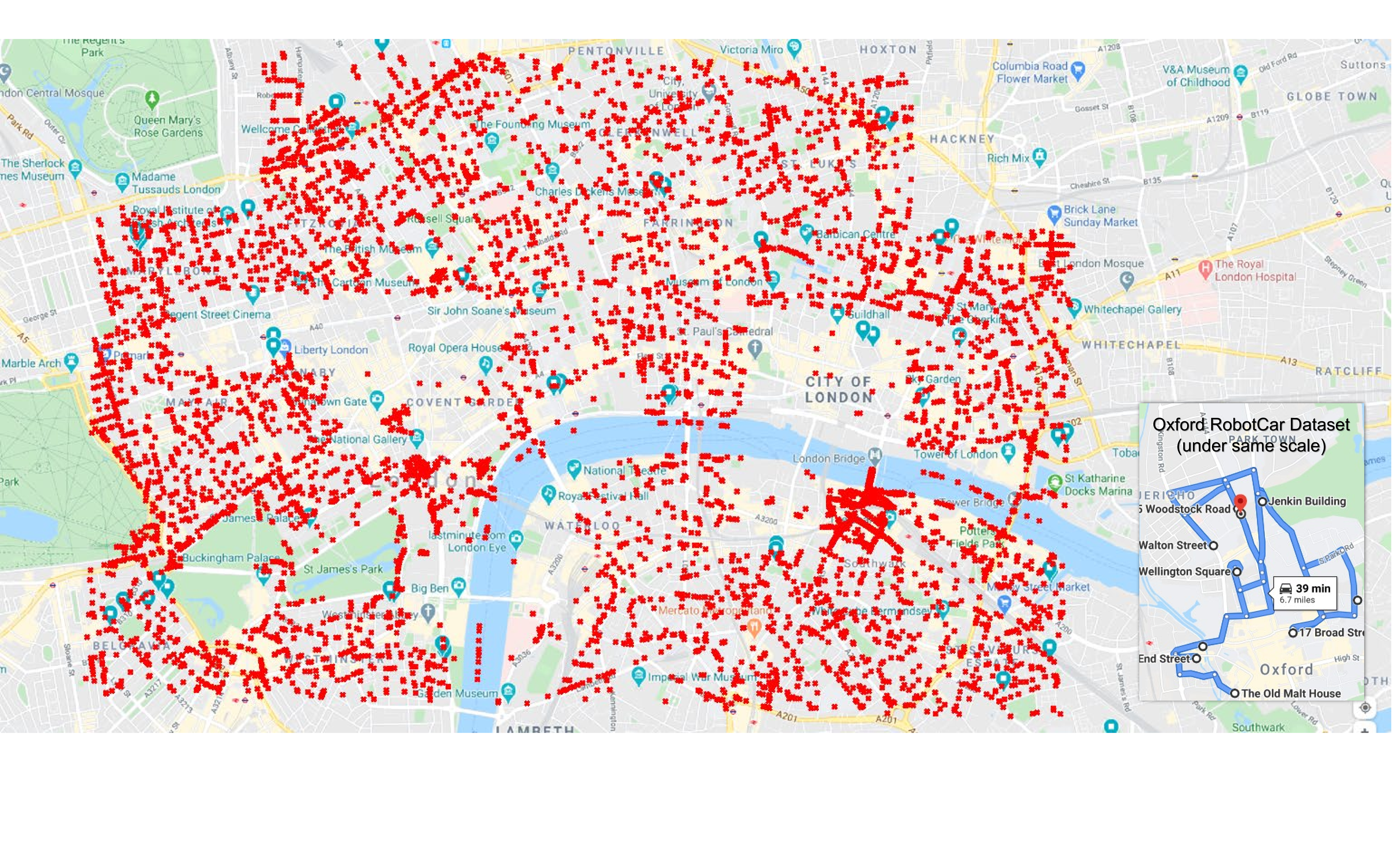}%
  }
  \vspace{-2.5mm}
  \subfloat[Panorama\label{fig:teaser:pano}]{%
    \includegraphics[height=.165\linewidth]{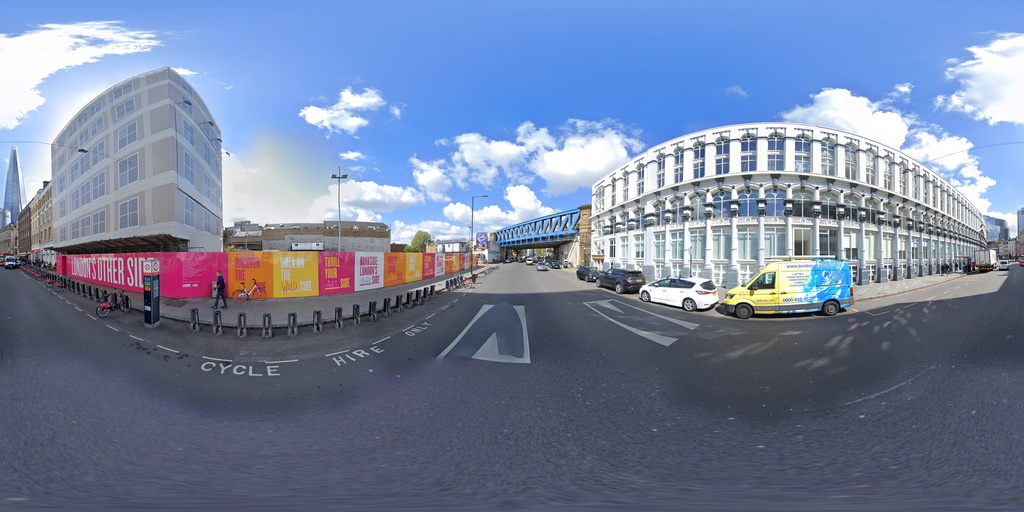}%
  }\hfill%
  \subfloat[{RGB}\label{fig:teaser:image}]{%
    \includegraphics[height=.165\linewidth]{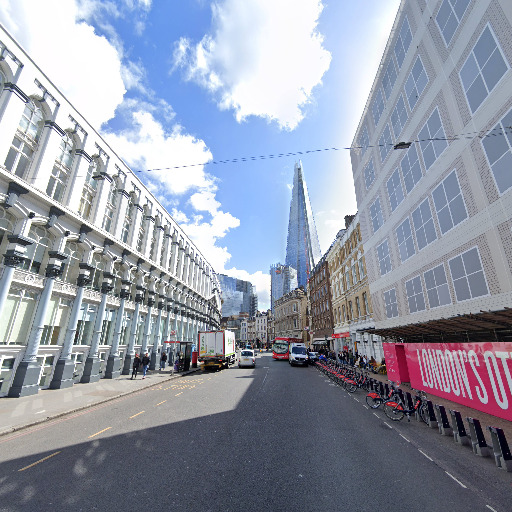}%
  }\hfill%
  \subfloat[Renderings (surface segments, depth, normal)\label{fig:teaser:renderings}]{%
    \includegraphics[height=.165\linewidth]{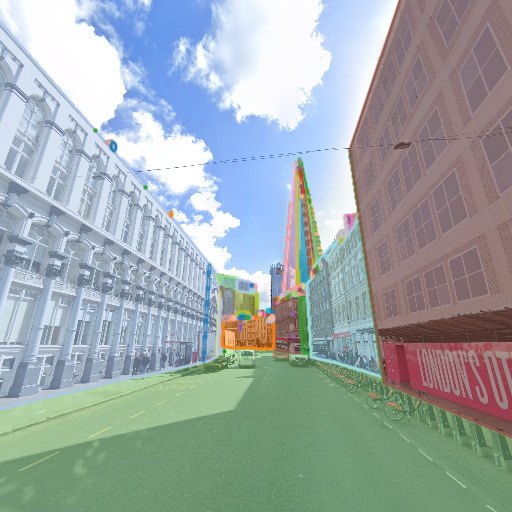}%
    \includegraphics[height=.165\linewidth]{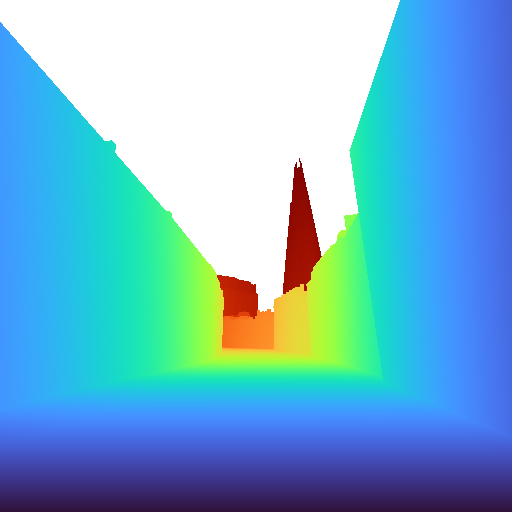}%
    \includegraphics[height=.165\linewidth]{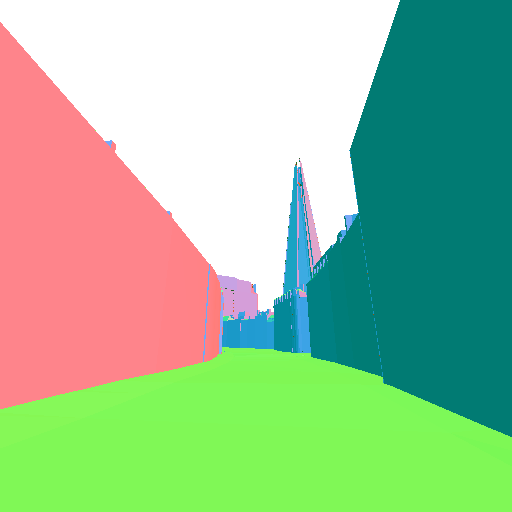}%
  }
  \caption{
    Our \OURS{} dataset consists of accurate city-scale CAD models and spatially-registered street
    view panoramas. \OURS{} covers an area of more than 20 km$^2$ in London from 6,300 viewpoints, which
    dwarfs previous datasets such as Oxford RobotCar \cite{maddern20171} (\ref{fig:teaser:satellite}).
    From the CAD models (\ref{fig:teaser:CAD}) and the panoramas
    (\ref{fig:teaser:pano}), it is possible to generate \simplify{all kinds of} clean structured 
    ground-truths for 3D understanding tasks, including perspective RGB images
    (\ref{fig:teaser:image}), surface segments, and normal maps (\ref{fig:teaser:renderings}).
  }
  \label{fig:teaser}
\end{figure*}

\begin{figure*}[t]
\centering
\def\imw{.165\linewidth}
\setlength{\tabcolsep}{0.5pt}
\begin{tabular}{cccccc}
\includegraphics[width=\imw]{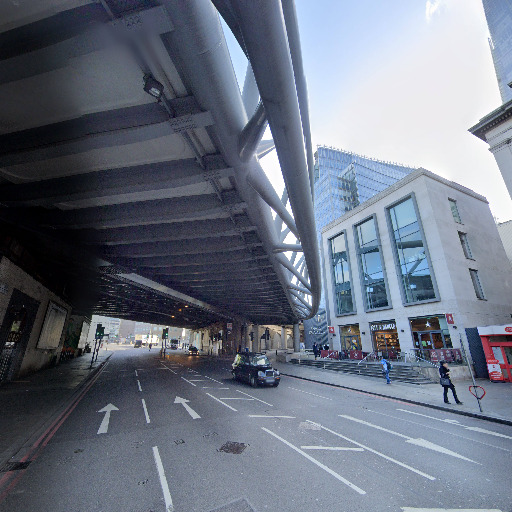} & 
\includegraphics[width=\imw]{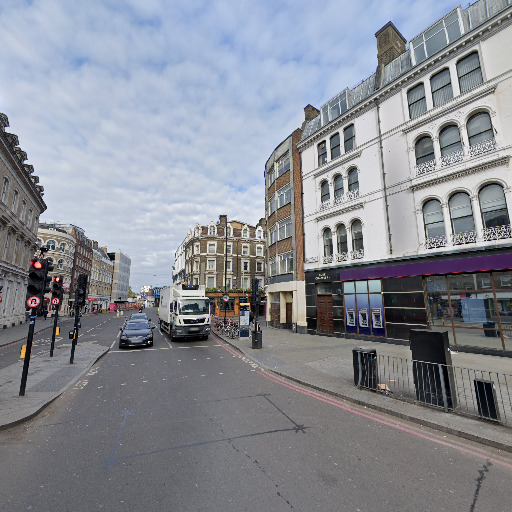} &
\includegraphics[width=\imw]{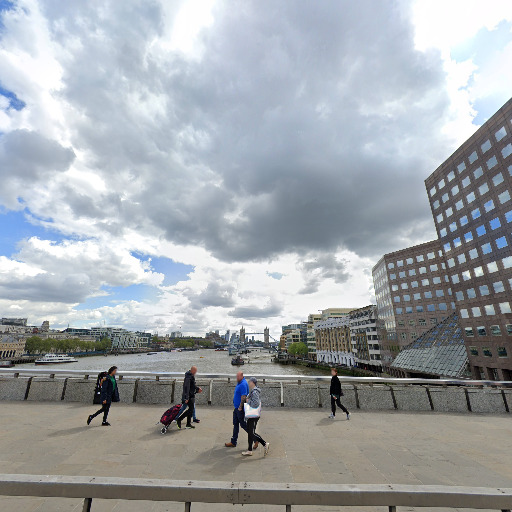} &
\includegraphics[width=\imw]{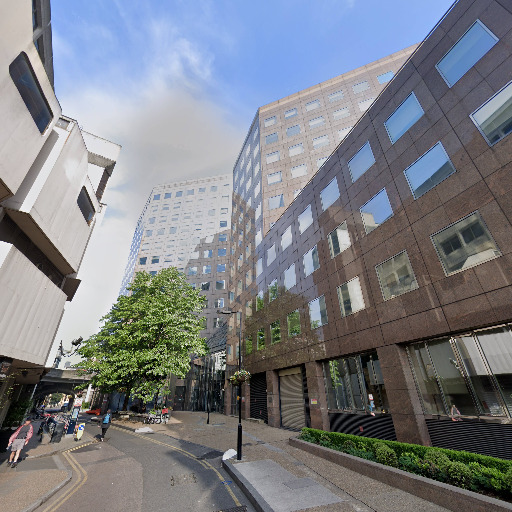} &
\includegraphics[width=\imw]{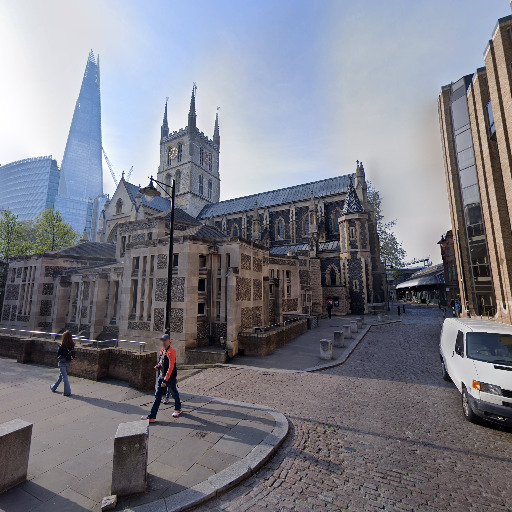} &
\includegraphics[width=\imw]{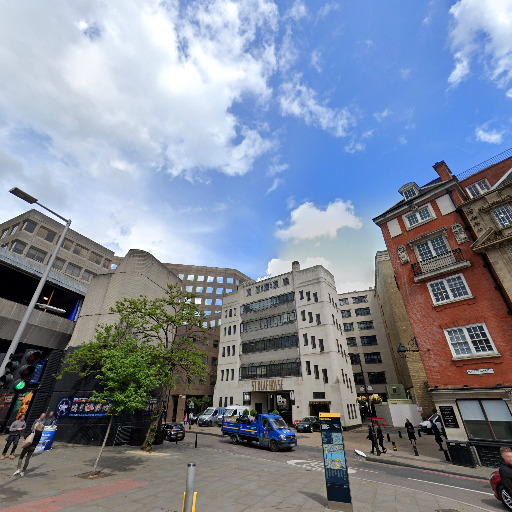} \\ [-2.5pt]
\includegraphics[width=\imw]{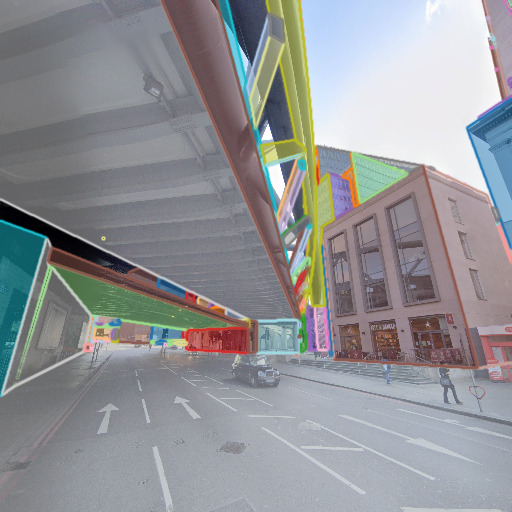} & 
\includegraphics[width=\imw]{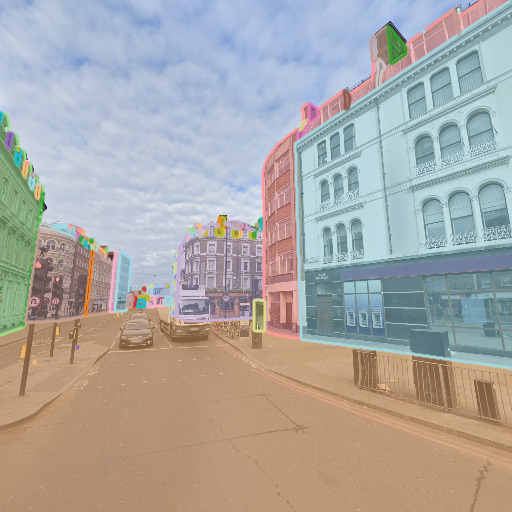} &
\includegraphics[width=\imw]{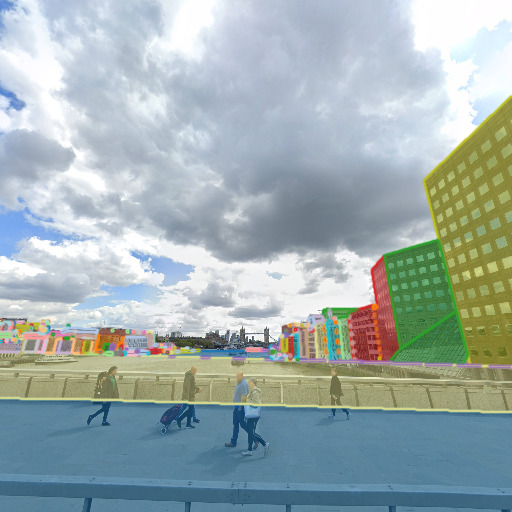} &
\includegraphics[width=\imw]{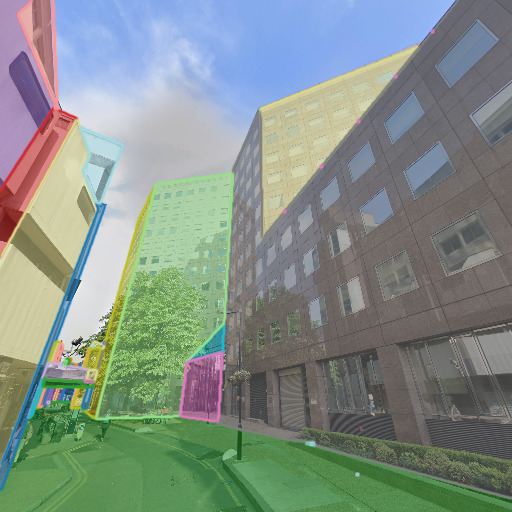} &
\includegraphics[width=\imw]{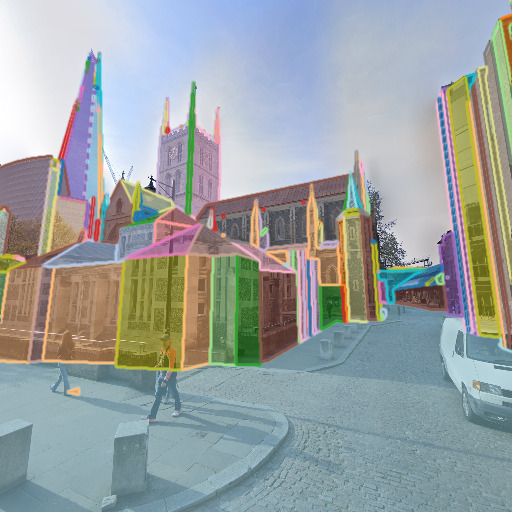} &
\includegraphics[width=\imw]{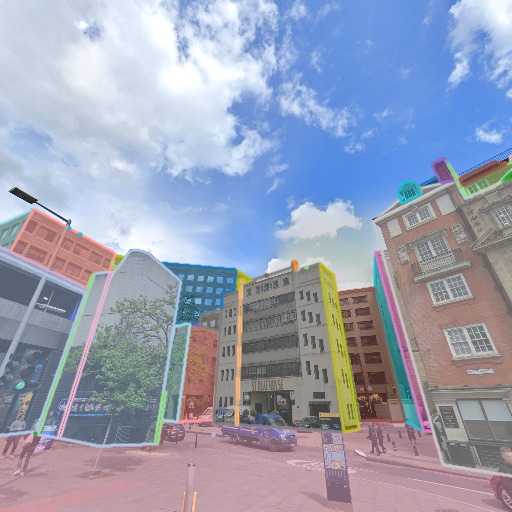} \\ [-2.5pt]
\includegraphics[width=\imw]{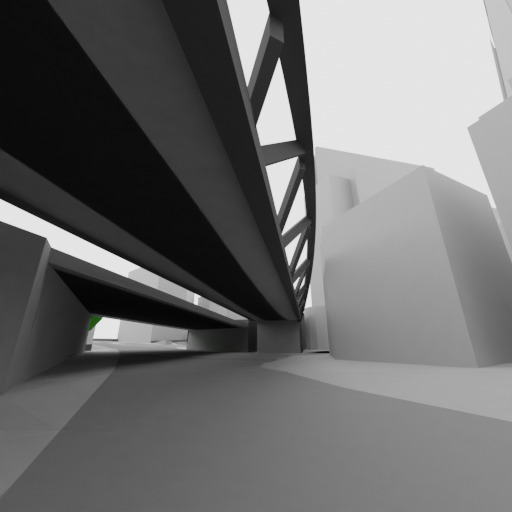} & 
\includegraphics[width=\imw]{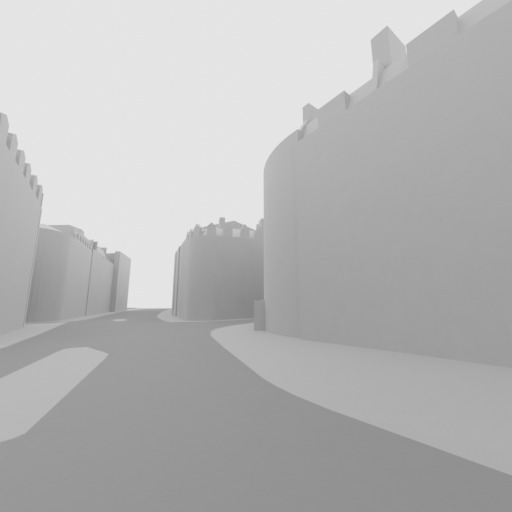} &
\includegraphics[width=\imw]{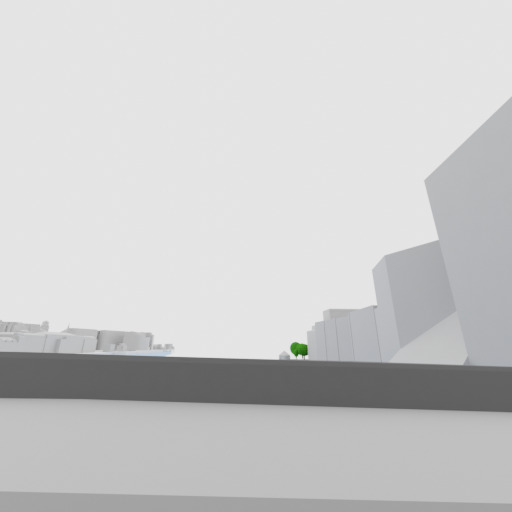} &
\includegraphics[width=\imw]{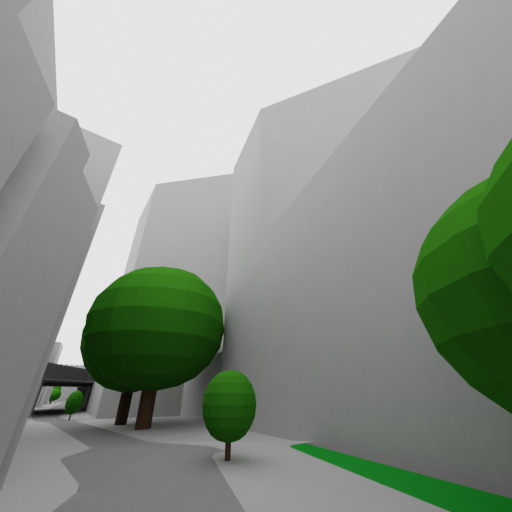} &
\includegraphics[width=\imw]{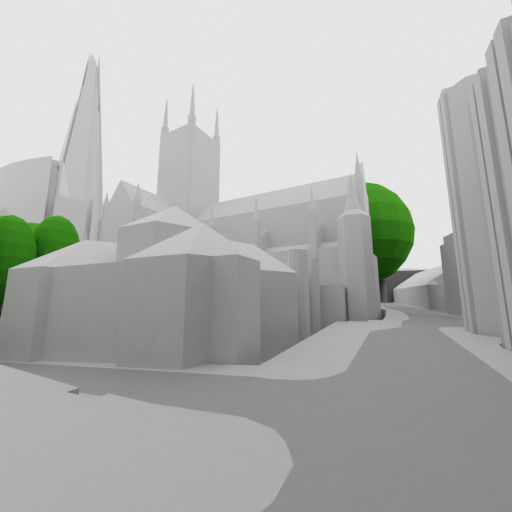} &
\includegraphics[width=\imw]{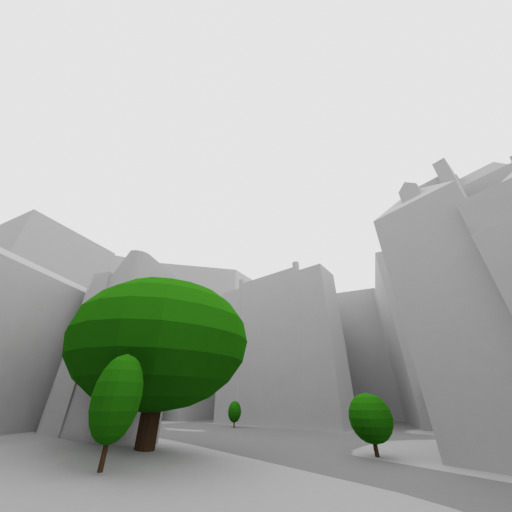} \\ [-2.5pt]
\includegraphics[width=\imw]{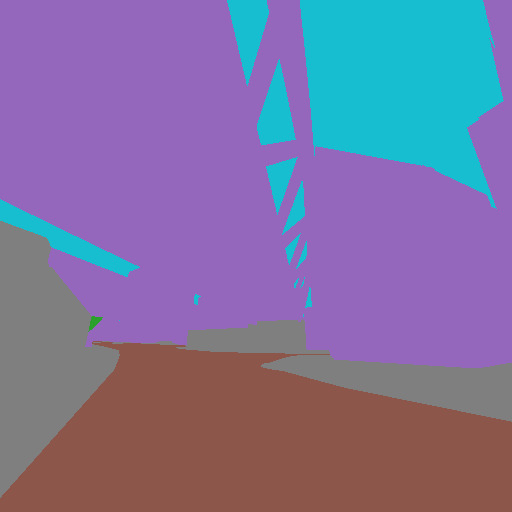} & 
\includegraphics[width=\imw]{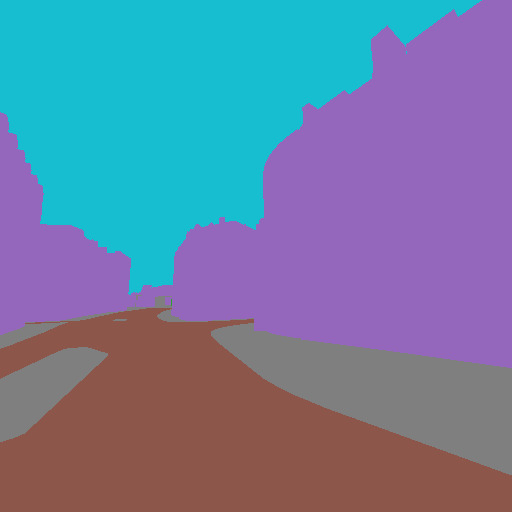} &
\includegraphics[width=\imw]{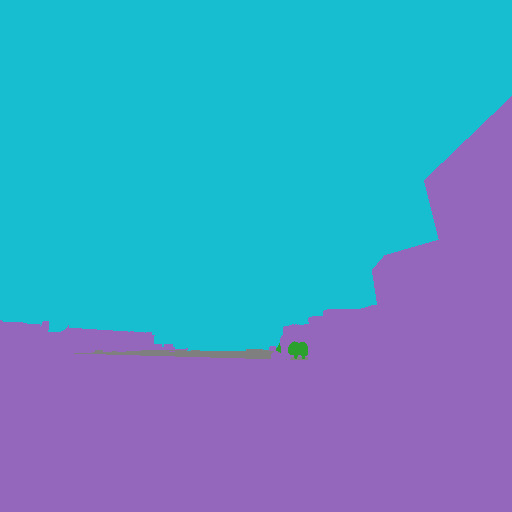} &
\includegraphics[width=\imw]{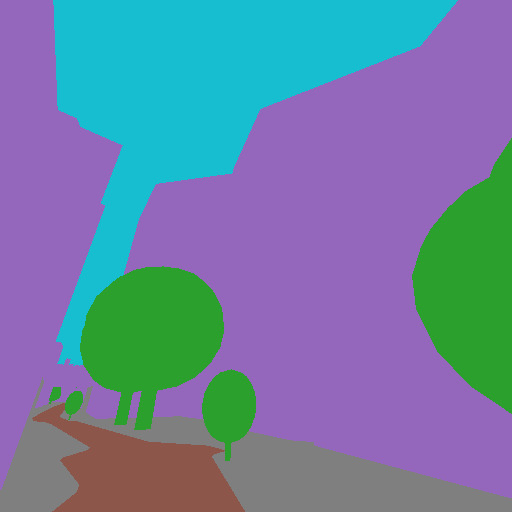} &
\includegraphics[width=\imw]{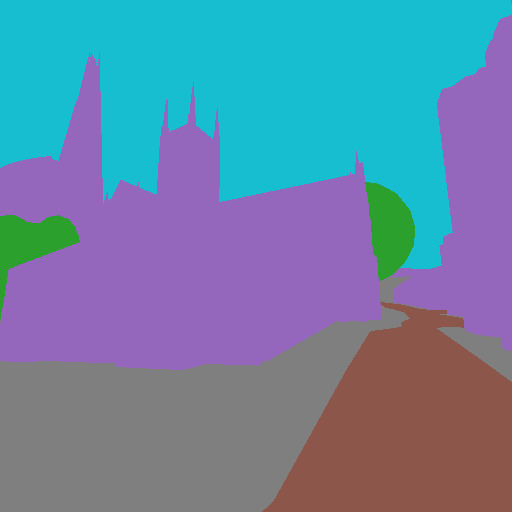} &
\includegraphics[width=\imw]{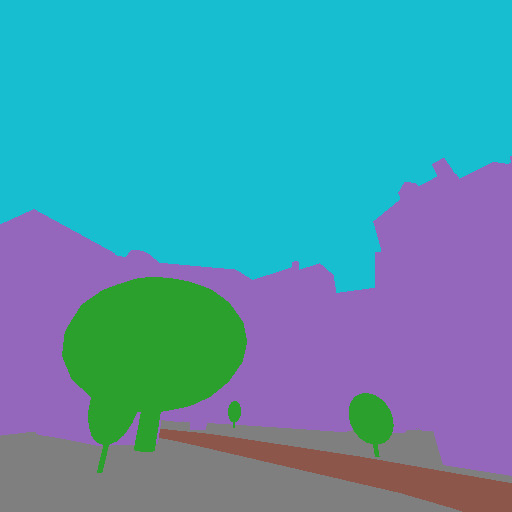} \\ [-2.5pt]
\end{tabular}
\caption{
  Images and generated 3D information from sampled viewpoints of \OURS{} dataset. From top to bottom: perspective images rendered from panoramas, surface segments overlaid with images, CAD model renderings, and semantic segmentation.
  }
 \label{fig:align}
\end{figure*}

\section{Introduction}

In the past decades, we have witnessed an increasing demand for 3D vision technologies.
With the development of point features such as ORB \cite{rublee2011orb} and SIFT \cite{arandjelovic2012three}, structure-from-motion (SfM) and simultaneous localization and mapping (SLAM) have been successfully applied to tasks such as autonomous driving, robotics, and augmented reality.
3D vision products, such as Hololens and Magic Leap can use them to reconstruct the environment.
Although the robustness of SfM has been improved over the last decades, the resulting point clouds are still noisy, incomplete, and thus can hardly be used directly.
Intricate post-processing procedures, such as plane fitting \cite{huang20173dlite}, Poisson surface reconstruction \cite{kazhdan2006poisson}, and TSDF fusion \cite{curless1996volumetric} are necessary for downstream applications.
Increasingly have people found that the long pipeline of 3D reconstruction is difficult to implement correctly and efficiently, and results in low-level representations such as point clouds are also unfriendly for parsing, editing, and sharing.

Looking back at the origin of computer vision from the '80s, scientists have found that human beings do not perceive the world with point features.
Instead, we abstract scenes with high-level geometry primitives, such as corners, line segments, and planes, to form our sense of 3D, navigate in cities and interact with environments \cite{clowes1971seeing}.
This reminds us that instead of point clouds, we can also use high-level structures as a representation for 3D reconstruction, which in many cases are more compact, intuitive, and easy to process.
In fact, early vision research does focus on reconstructing with high-level abstractions, such as lines/wireframes \cite{falk1972interpretation}, contours/boundaries \cite{stevens1981visual}, planes/surfaces \cite{kanade1981recovery}, and cuboids/polyhedrons \cite{roberts1963machine}.
We name these high-level abstractions \emph{holistic structures} in this paper, as they tend to represent scenes globally, comparing to the SIFT-like local features.
However, recognition of holistic structures from images seems too challenging to be practical at that time. 3D reconstruction with high-level abstractions does not get enough attention despite its potentials, until recently.

Inspired by the recent success of deep convolutional neural networks, researchers have proposed a variety of neural networks to extract high-level structures from images, such as wireframes \cite{zhou2019learning,zhou2019end}, planes \cite{liu2018planenet}, cuboids \cite{mousavian20173d}, vanishing points \cite{zhou2019neurvps}, room layouts \cite{zou2018layoutnet}, and building layouts \cite{zeng2020bundle}.
Most of them are supervised learning algorithms, which rely on annotated datasets for training.
However, making a properly outdoor 3D dataset for a particular high-level representation is complex.
The building process usually has 2 stages: (1) 3D data collection and (2) structure labeling.
Collecting 3D data such as depth images is a cost- and labor-intensive process.
This is especially true for outdoor scenes due to the lack of dense depth sensors.
Even with expensive LiDAR systems, the point clouds from scans are still noisy and have lower spatial resolution compared to RGB images.
Derived features such as surface normal are unsmooth, which might be the reason why previous normal estimation research \cite{eigen2015predicting,bansal2016marr,wang2015designing,huang2019framenet} only demonstrate their results on indoor scenes.
These characteristics are unfavorable for extracting holistic structures.
Besides, labeling high-level abstractions on the collected 3D data is also challenging.
On one hand, manually annotating high-level structures is time-consuming, as it requires researchers to design complicated 3D labeling software.
On the other hand, the quality of automatically extracted structures from algorithms such as J-Linkage \cite{toldo2008robust} might not be adequate.
The results can be inaccurate, incomplete, and erroneous, especially when the quality of 3D data is not that good\simplify{ and the supporting features of the high-level structure are small}.
To make the problem worse, frequently a dataset that is labeled for one particular structure cannot be easily reused for other structures.
As a result, data preparation has become one of the major road blockers for structural 3D vision research.

To \simplify{address the aforementioned challenges and} provide a high-quality multi-purpose dataset for the vision community, we develop  ``HoliCity,'' a data platform for learning holistic 3D structures in urban environments.
\Cref{fig:teaser} shows an overview.
HoliCity is composed of 6,300 high-resolution real-world panoramas that are accurately aligned with the 3D CAD model of downtown London with more than 20 km$^2$ of area (see \Cref{fig:teaser:CAD,fig:teaser:satellite,fig:teaser:pano}).
Instead of relying on expensive vehicle-mounted LiDAR scanners, HoliCity takes the advantage of existing high-quality 3D CAD city models from the GIS community.
This way, we can collect a large area of 3D data with fine details, structure-level annotation, and semantic labels at relatively low cost, in which the CAD models are parametrized by corners, lines, and smooth surfaces so that it is friendly for researchers to extract holistic structures.
In comparison, traditional LiDAR-based datasets such as KITTI \cite{geiger2013vision} and RobotCar \cite{maddern20171} cover a much smaller area (see \Cref{fig:teaser:satellite} for comparison), are more expensive to collect, and use noisy point clouds as their representation. The panorama photographs in HoliCity are sharp, professionally captured, and with resolution as high as $13312 \times 6656$.
In contrast, images of LiDAR-based datasets are often from video recordings, so they can be blurry, low-resolution, and repetitive.
Application-wise, traditional LiDAR-based datasets focus on tasks related to low-level representations, such as depth map prediction, reconstruction with point clouds, and camera relocalization,
while HoliCity is designed to additionally support the research of 3D reconstruction with high-level holistic structures, such as junctions, lines, wireframes, planes, parameterized surfaces, and other geometry primitives that can be derived from CAD models.

In summary, the main contributions of this work include:
\begin{enumerate}
  \item we propose a novel pipeline for creating a city-scale 3D dataset by utilizing existing CAD models and street-view imagery at a relatively low cost;
  \item we develop HoliCity as a data platform for learning holistic 3D structures in urban environments;
  \item we accurately align the panorama images with the CAD models, in which the median reprojection error is less than half a degree for an average image;
  \item we conduct experiments to justify the necessity of a CAD model-based data platform for 3D vision research, including demonstrating potential applications and testing its generalizability from/to other datasets.
\end{enumerate}

\begin{table*}[t]
\centering
\begin{small}
\resizebox{\textwidth}{!}{%
\begin{tabular}{c|cccccccc}
              & NYUv2                          & ScanNet                        & Stanford-2D-3D & SYNTHIA                         & MegaDepth  & KITTI                          & RobotCar   & \textbf{\OURS}      \\ \hline
type          & real                           & real                           & real           & synthetic                       & real       & real                           & real       & \textbf{real}       \\
scene         & indoor                         & indoor                         & indoor         & driving                         & landmark   & driving                        & driving    & \textbf{city}       \\
depth         & RGBD                           & RGBD                           & RGBD           & CAD                             & SfM        & LIDAR                          & LIDAR      & \textbf{CAD}        \\
style         & dense                          & dense                          & $\bigcirc$     & dense                           & dense      & quasi                          & quasi      & \textbf{dense}      \\
normal        & $\bigcirc$                     & \checkmark                     & \checkmark     & $\bigcirc$                      & $\bigcirc$ & $\bigcirc$                     & $\bigcirc$ & \textbf{\checkmark} \\
plane         & $\bigcirc$                     & \checkmark                   & $\bigcirc$     & \checkmark                      & $\bigcirc$ & $\bigcirc$                     & $\bigcirc$ & \textbf{\checkmark} \\
coverage      & /                              & 0.034 km$^2$                   & 0.006 km$^2$   & /                               & /          & /                              & /          & \textbf{20 km$^2$}  \\
path length   & /                              & /                              & /              & /                               & /          & 39.2 km                        & 10 km      & \textbf{/}          \\
time span     & 1 scan                         & 1 scan                         & 1 scan         & /                               & unknown    & 5 scans                        & 2014-2015  & \textbf{2008-2019}  \\
diversity     & 464 rooms                      & 707 rooms                      & 4 buildings    & /                               & 200 scenes & path                           & path       & \textbf{city}       \\
\# of  images & 1.4k                           & 2.5m                           & 1.4k           & 50k                             & 100k       & 93k                            & 20m        & \textbf{6.3k}       \\
source        & image                          & video                          & video          & /                               & image      & video                          & video      & \textbf{image}      \\
FoVs          & $71^{\circ} \times 60^{\circ}$ & $45^{\circ} \times 34^{\circ}$ & panorama       & $100^{\circ} \times 84^{\circ}$ & random     & $90^{\circ} \times 35^{\circ}$ & multi-cam  & \textbf{panorama}   \\
sementics     & 2D                             & 3D                             & 3D             & 3D                              & N.A.   & N.A.                       & N.A.   & \textbf{3D}         \\
max depth     & (indoor)                       & (indoor)                       & (indoor)       & $\infty$                        & (relative) & 80m                            & 50m        & \textbf{$\infty$}  
\end{tabular}%
}
\end{small}
  \caption{Comparing HoliCity with existing 3D datasets. We list the features of NYUv2 \cite{Silberman:ECCV12}, ScanNet\cite{dai2017scannet},  Stanford-2D-3D-Semantics \cite{armeni2017joint}, SYNTHIA \cite{ros2016synthia}, MegaDepth \cite{zhengqi2018megadepth}, KITTI \cite{geiger2013vision}, and RobotCar \cite{maddern20171}.  The $\bigcirc$ in the normal and plane rows represents that they are not available but possible to use fitting algorithms such as J-Linkage \cite{toldo2008robust} to get the annotations, but the quality might vary\simplify{due to the noise in point clouds}.} 
\label{tab:dataset}
\end{table*}

\section{Related Work} \label{sec:related}

\paragraph{Synthetic 3D Datasets.}
Recently, object-level synthetic datasets such as ShapeNet \cite{chang2015shapenet} are popular for computer vision research, as people are free to convert 3D CAD models to any representations that their learning-based algorithms like, such as depth maps \cite{chang2018pyramid}, meshes \cite{mescheder2019occupancy}, voxels \cite{yan2016perspective}, point clouds \cite{fan2017point}, and signed distance fields \cite{park2019deepsdf}.
With the availability of CAD models, not only HoliCity shares similar freedom as these synthetic object-level datasets, but it also offers scene-level real-world images in urban environments.
Additionally, synthetic approaches have also been used to create structured 3D scenes, as seen in SceneNet \cite{mccormac2017scenenet}, SUNCG \cite{song2016ssc}, SYNTHIA \cite{ros2016synthia} and GTA5 \cite{richter2016playing} datasets.
Nevertheless, their images are still fake.
In our experiments, we find that there exists a large domain gap between the virtual renderings of synthetic datasets and our real-world images.

\paragraph{Outdoor Datasets.}
Due to the high cost and the limitation of LiDAR systems, acquiring 3D measurements for outdoor scenes is difficult.
Publicly available datasets created with LiDAR technology, such as KITTI \cite{geiger2013vision} and RobotCar \cite{maddern20171}, are relatively small-scale and low-resolution, and mainly focus on the driving scenarios\simplify{where the camera is facing toward the road}.
Recently, outdoor datasets emerge by leveraging structure-from-motion (SfM) and multi-view stereo (MVS) on web imagery in-the-wild \cite{zhengqi2018megadepth}.
These datasets provide depth information at a low cost with the expense of quality, because visual 3D reconstruction is not accurate or robust for random Internet images.
In addition, previous 3D outdoor datasets mainly use point clouds as their representation, which are usually noisy.
Hardly any of them provide structured annotations such as lines, wireframes, segmented 3D planes, and buildings instance.
In comparison, HoliCity offers high-quality CAD models and ground truth of holistic 3D structures that cover an unprecedented range of areas and viewpoints at the scale of a city (\Cref{fig:teaser,fig:align}).

\paragraph{Indoor Datasets.}
Thanks to increasingly affordable indoor dense depth sensors such as Kinect and RealSense, high-quality real-world indoor 3D data can be produced on a massive scale.
Datasets like NYUv2 \cite{Silberman:ECCV12} provide RGBD images for a variety of indoor scenes.
Recent datasets such as SUN3D \cite{xiao2013sun3d}, ScanNet \cite{dai2017scannet}, Stanford-2D-3D-Semantics \cite{armeni2017joint}, and Matterport3D \cite{Matterport3D} provide surface reconstruction results and 3D sementics annotation in addition to depth maps.
The quality of indoor datasets often varies from scene to scene, depending on how well the scene is scanned.
Compared to HoliCity that provides accurate CAD models designed for learning holistic structures, the noises, holes, and misalignments in the point clouds of these indoor datasets make them not ideal for extracting high-level 3D abstractions.
More importantly, our experiment shows that it is hard for a network to generalize from indoor training data to outdoor 3D tasks, due to the significant domain gaps.

\begin{figure*}[t]
\vspace{-0.18in}
\centering
\subfloat[Annotations per Image \label{fig:statistics:count} ]{\includegraphics[width=.33\linewidth]{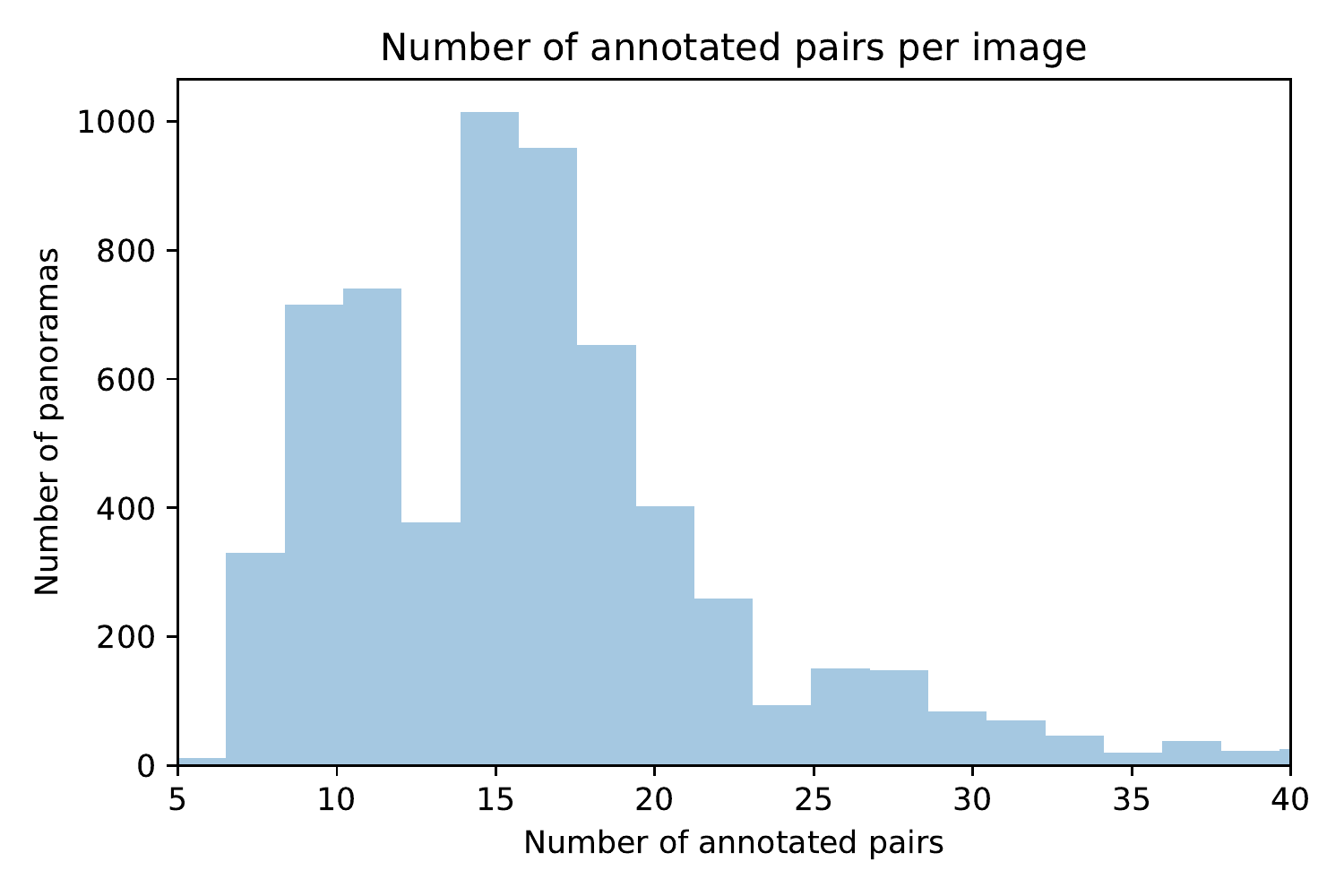}}%
\subfloat[Reprojection Error \label{fig:statistics:reprojection} ]{\includegraphics[width=.33\linewidth]{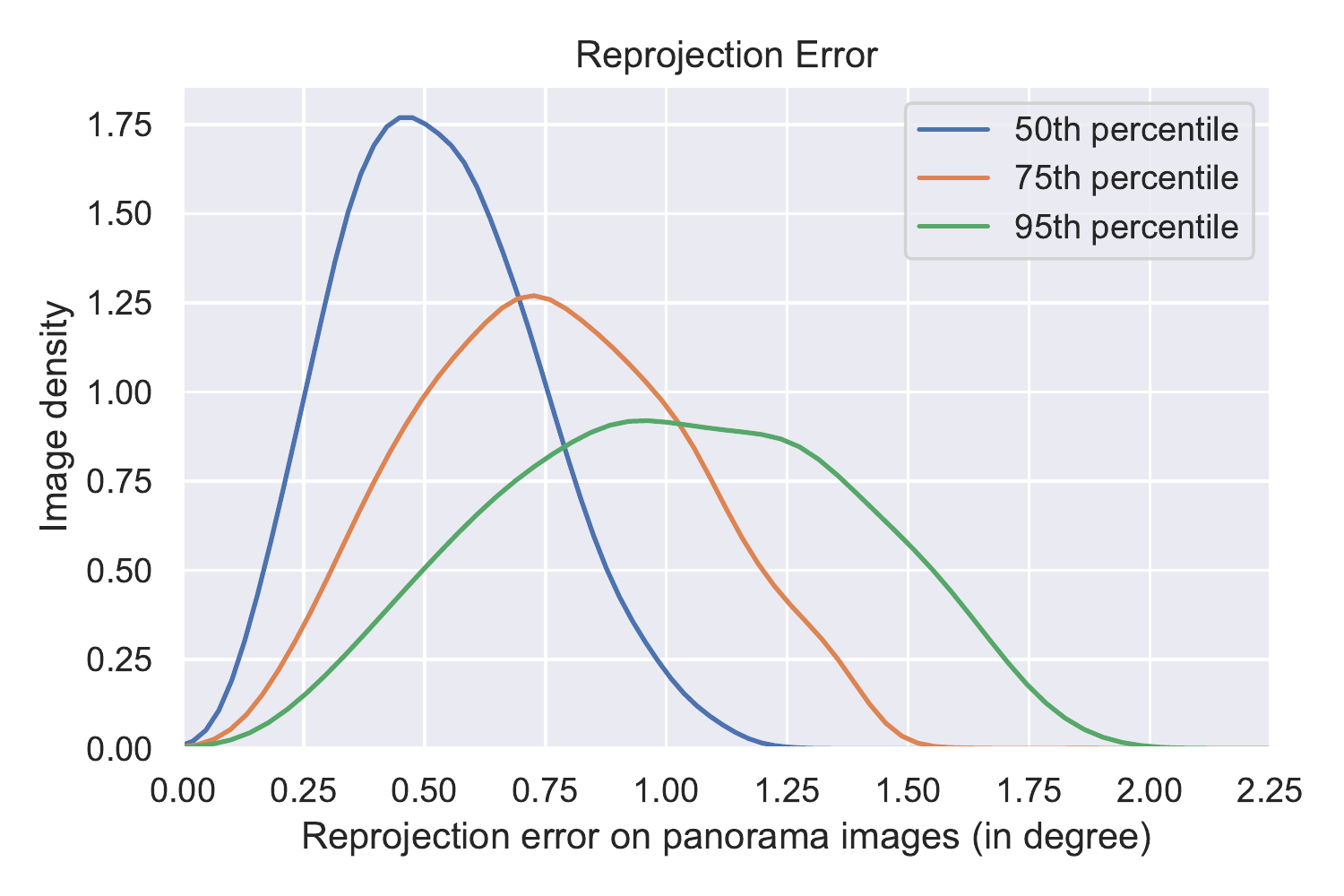}}%
\subfloat[Plane Occurrence \label{fig:statistics:viewers} ]{\includegraphics[width=.33\linewidth]{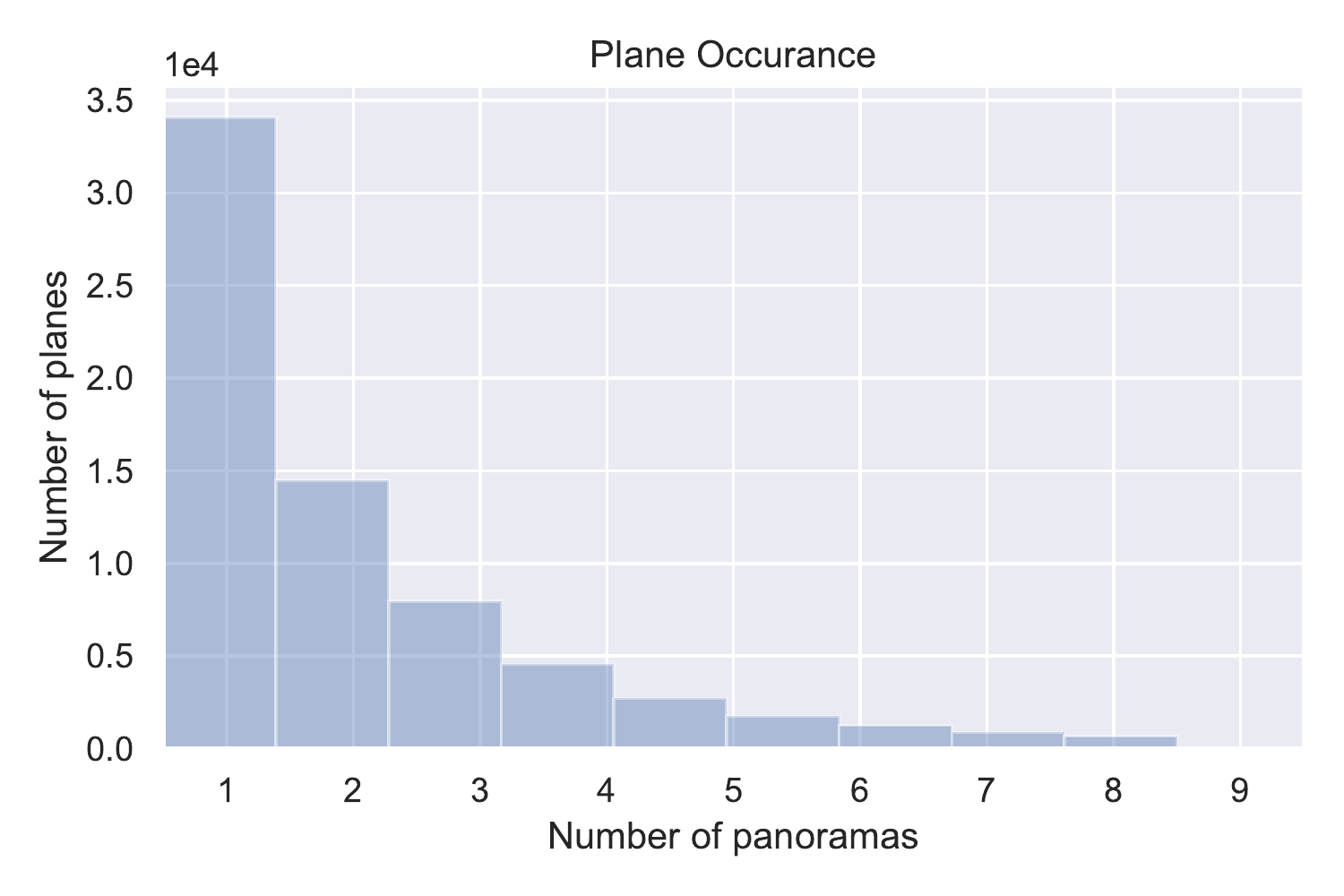}}

\caption{Statistics of \OURS{}. We show the number of annotations per panorama for registration (\ref{fig:statistics:count}); the reprojection errors of annotated 3D points on panoramas (\ref{fig:statistics:reprojection}) and the occurrence of planes on different panoramas (\ref{fig:statistics:viewers}).
\label{fig:statistics}}
\end{figure*}

\section{Exploring \OURS} \label{sec:exploring}

Our goal is to develop a large-scale outdoor 3D dataset that is rich in holistic structural information. To this end, \OURS{} uses commercially available CAD models provided by AccuCities\simplify{\footnote{https://www.accucities.com/}}, which are reconstructed and built using photogrammetry from high-resolution aerial imagery, each with accurately-recorded GPS position, height, tilt, pitch, and roll. Aerial photogrammetry is a mature technique and it has been widely used to build models with different levels of details for city planning in the field of geographic information systems (GIS). As a result, we are able to get CAD models that cover a wide range of city areas. The CAD models we use contain details of building features with up to 15 cm accuracy, according to the provider.

To make the CAD models useful for image-based tasks, we need to precisely align the CAD models with images taken from the ground. We collect the panorama images from Google Street View that cover the same area.  To increase the alignment accuracy between the panoramas and the CAD models, we implement an annotation software for precise localization (\Cref{sec:building}).

\Cref{tab:dataset} summarizes the difference between our dataset and the existing ones. Compared to the previous 3D outdoor datasets, \OURS{} has advantage in the following aspects:

\paragraph{Holistic Structures.} 
With CAD models, it is straightforward to extract high-quality holistic structures such as corners, lines, planes, and even curved surfaces from HoliCity compared to the point clouds, as shown in the second row of \Cref{fig:align}.
HoliCity also supports traditional low-level 3D representations such as depth maps and normal maps (\Cref{fig:teaser:renderings}), as well as some 2D tasks because we provide rendering semantics renderings of roads, buildings, curbs, sky, water, and others. 
In contrast, most existing outdoor datasets use point clouds as the representation.
Due to the limitation of LiDAR technology and costs, point clouds are often too sparse and noisy to generate such high-level structures reliably.
Hardly any existing outdoor datasets provide high-level structure annotations such as lines, wireframes, segmented surfaces, and identified buildings.

\paragraph{Coverage.}
Compared to other datasets, HoliCity can cover a much larger area of more than 20 km${}^2$ in downtown London with more diverse urban scenes and viewpoints, thanks to the existing CAD models and street-view panoramas.
\Cref{fig:teaser:satellite} shows the coverage map that is aligned with Google Maps and compares it against the Oxford RobotCar dataset.
HoliCity contains 6,300 panorama images from diverse viewpoints.
We note that it might look like that our dataset has fewer images than other datasets, it is large among panorama-based ones, such as Stanfold-2D-3D \cite{armeni2017joint} (1,413 images), and SUN3D \cite{xiao2013sun3d} (6,161 images).
For the datasets in \Cref{tab:dataset} with much higher image counts, their images are mostly extracted from videos, which are highly repetitive and blurry.
We believe ``coverage'' is a more fair metric for evaluating the size and variety of a dataset, especially considering that our dataset already has had a reasonable density of viewpoints as seen in \Cref{fig:teaser:satellite}.

\paragraph{Accuracy.}
We carefully align the panoramas with the CAD model using a reasonable number of annotated correspondence points between them, as shown in \Cref{fig:statistics:count}.
Because the original geolocation of Google Street View images is not precise, we re-estimate the camera pose by minimizing the reprojection error of our annotations.
\Cref{fig:statistics:reprojection} shows the reprojection error of annotated points between the images and the CAD model.
We find that for an average image, the median reprojection error is less than half a degree and the 95th percentile does not exceed 1.2 degrees.
Besides accurate camera registration, our CAD model-based dataset does not have constraints on maximum depth, unlike the depth obtained from LiDAR.
Hence it is more suitable for evaluating image-based 3D reconstruction algorithms.

\paragraph{Panorama.}
HoliCity uses panorama images from Google Street View with a resolution $13312 \times 6656$.
This way, our dataset can capture the full view from each viewpoint and it is not biased towards any directions or landmarks.
It also gives us extra flexibility to render many times more perspective images and emulate cameras of different types.
In contrast, images in previous outdoor datasets are mainly captured by the front-facing cameras, in which the direction is biased and field of views is limited.

\paragraph{Multi-View.}
The number of occurrences of each 3D plane in our panorama database is shown in \Cref{fig:statistics:viewers}.
More than half of the planes occur in more than one image and about a third of planes occur in more than two images.
This means that our dataset can potentially support the 3D vision project that uses multi-view correspondence with large baselines, e.g., SfM, MVS, and neural renderings.

\paragraph{Time Span.}
Most of existing 3D outdoor datasets are collected in short periods, as shown in the row ``time span'' of \Cref{tab:dataset}.
In contrast, HoliCity utilizes the panorama images from Google Street View, which are captured during a span of over 10 years.
This greatly increases the variety of data, which can benefit learning-based methods and bring additional challenges to tasks.

\section{Building \OURS}
\label{sec:building}

\subsection{City Data Collection} 
\paragraph{3D Models.} Although there exist many public city CAD models from the GIS community \cite{kolbe2005citygml} and municipality governments\simplify{\footnote{Related resources are summarized at \url{https://3d.bk.tudelft.nl/opendata/opencities}.}}, determining their quality is hard as these datasets are built for different purposes. In this project, we use the commercially available CAD model from AccuCities.  Their CAD model covers the area of downtown London and comes with two levels of details. The low-resolution version (cover 20 km${}^2$) has details accurate to 2m, while the high-resolution version (covers 4 km${}^2$) is accurate to 15cm in all three axes.  The CAD model is stored in the mesh format and each surface is tagged with semantic types such as \texttt{BUILDING}, \texttt{TERRAIN}, \texttt{BRIDGE}, \texttt{TREE}, etc.

\paragraph{Street-View Images.} We collect street-view panorama images from Google Street View.  At each viewpoint, we have a 360$^{\circ}$ panorama image along with the geographic data of the camera from GPS and IMUs: 1) latitude and longitude in WGS84 coordinate; 2) azimuth, the angle between the forward-up plane of the camera and geographic north; 3) accelerating direction, which normally is the direction of gravity.
The geographic information along from Google Street View is not sufficient for accurately registering the camera pose between the CAD model and the panorama images. First, we do not have the elevation of the camera.  We estimate the initial $z$ of the camera by adding $2.5$m (the height of the camera) to the ground elevation, as terrains are provided in the CAD model. Second, the provided GPS and IMU data along are not accurate enough.  Therefore, we resort to human annotation for registration.

\subsection{Annotation Pipeline}
\simplify{Our annotation pipeline contains two steps: 1) registering the CAD model with the WGS84 coordinate by annotating key points on Google Maps and CAD models; 2) fine-tuning the registration by labeling the 2D-3D correspondence between the vertices of the CAD model and the pixels of the panorama.}

\paragraph{Geo-tagging the CAD Models.}
In the first step, we register the CAD model with the WGS84 coordinate used by Google Street View. To do that, we annotate 44 corresponding 2D locations on both Google Maps and our CAD model.  We label most points on the inner corners of roof ridges to maximize the registration accuracy.  We employ a nonlinear mesh deformation model.  Let $\Xwgs$ and $\Xcad$ be the 2D coordinates of the points on Google Maps and our CAD models and $\Gamma$ be the mapping from $\Xcad$ to $\Xwgs$ parameterized by $\Omega$.  Mathematically, we have
\begin{equation}
    \Gamma(\Xcad, \Omega)=\Xwgs 
\end{equation}
Here, we use $\Omega[x,y] \in \mathbb{R}^2$ is a 2D lookup table and $\Gamma$ simply bilinear interpolates $\Omega$ and returns $\Omega[\Xcad]$.  We can find the optimal $\hat\Omega$ by optimizing
\begin{equation}
    \min \|\Gamma(\Xcad, \Omega)-\Xwgs \|_2^2 + \lambda \|\Delta \Omega\|_F^2,
\end{equation}
where $\Delta \Omega$ is the Laplacian of $\Omega$.  The Laplacian term serves as a regularization.  The objective function is convex, so we can \simplify{solve it and} find the global optimal solution. We do 44-fold cross-validation to determine the best $\lambda$. The final average and maximum errors are 39cm and 1.5m in the cross-validation, respectively.  For reverse mapping from WGS84 to the CAD model, we use the Newton-Gaussian algorithm \simplify{to find the optimal $\Xcad$ that minimizes $\|\Gamma(\Xcad, \hat\Omega)=\Xwgs\|_2^2$}.

\paragraph{Per-Image Fine-Tuning.}
In the second step, we fine-tune the camera pose.  For each image, we ask annotators whether it is indoor or outdoor. We discard all the indoor images.  Next, we let annotators label pairs of corresponding points on the CAD model and the panoramas.
We instruct the annotator to only put points on roof corners if possible,  as our CAD model is made from aerial images so that the geometry of roofs is more reliable.

We apply Levenberg–Marquardt algorithm to compute the camera pose that minimizes the reprojection error.  Mathematically, let $\x_i \in \mathbb{S}^3$ be the unit vector of the ray direction of the $i$th labeled point on the panorama image and $\X_i \in \mathbb{R}^3$ be the coordinate of the corresponding vertex in the CAD world space.  The problem can be formulated as finding the best 6-DoF pose $\Theta$ (parameterized by its location, azimuth, and up direction) that minimizes the reprojection error:
\begin{equation}
  \min_{\Theta} \sum_{i=1}^n \arccos^2\left(\langle \x_i, \mathbf{P}_\Theta(\X_i) \rangle\right),
\end{equation}
where $\mathbf{P}_\Theta$ projects the world-space coordinate to the panorama space $\mathbb{S}^3$ with respect to the camera pose $\Theta$.

\begin{figure*}[t]
\setlength{\lineskip}{0.0ex}
\def\figsize{0.099\linewidth}
\centering

\includegraphics[width=\figsize,frame]{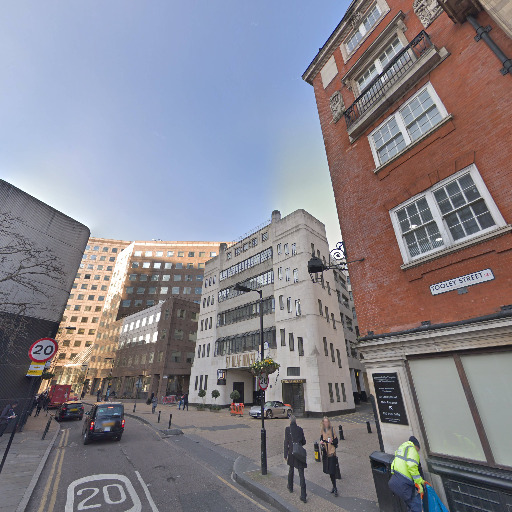}%
\includegraphics[width=\figsize,frame]{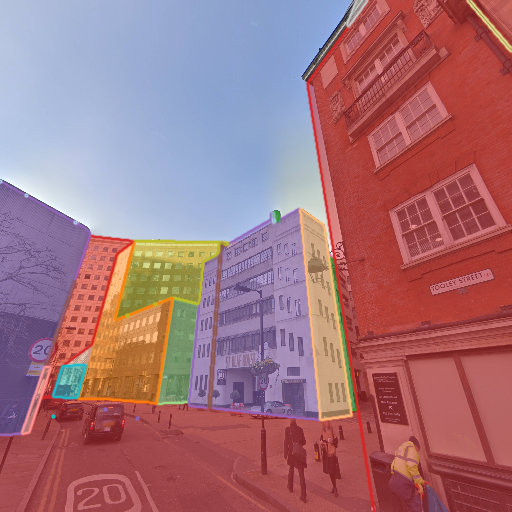}%
\includegraphics[width=\figsize,frame]{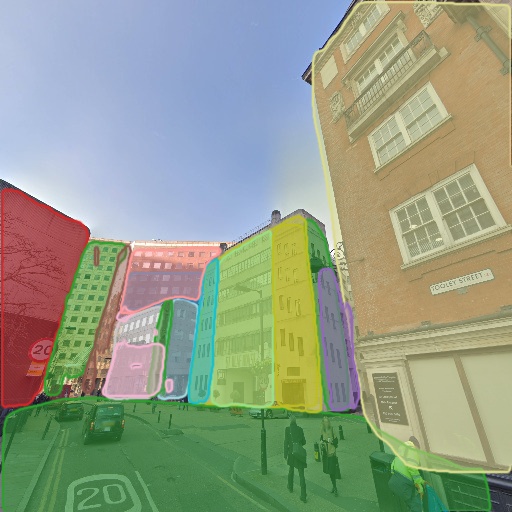}%
\includegraphics[width=\figsize,frame]{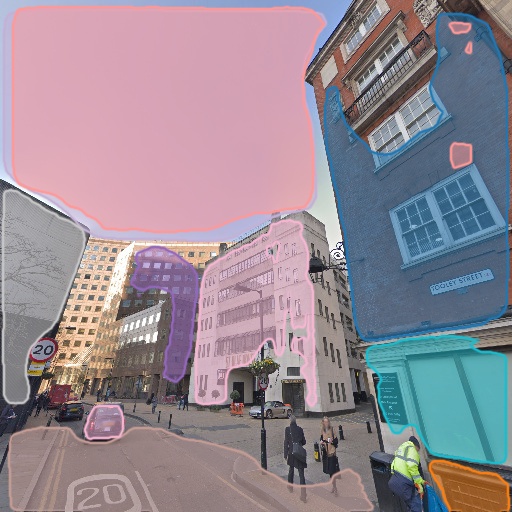}%
\includegraphics[width=\figsize,frame]{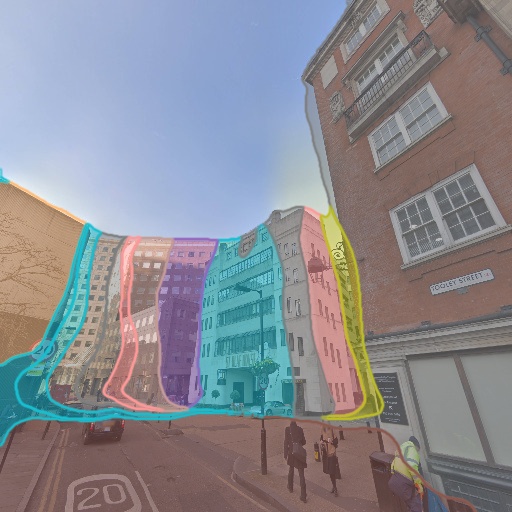}%
\includegraphics[width=\figsize,frame]{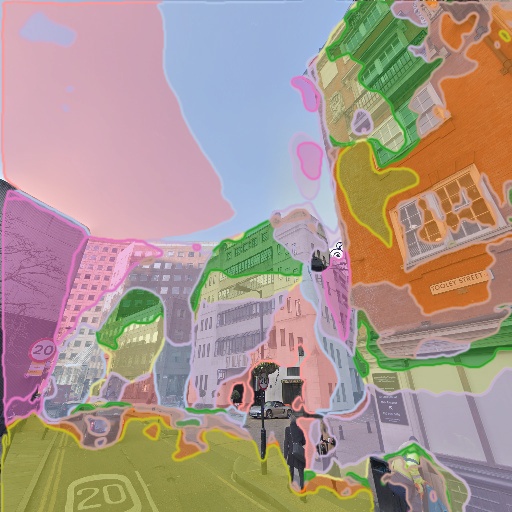}%
\includegraphics[width=\figsize,frame]{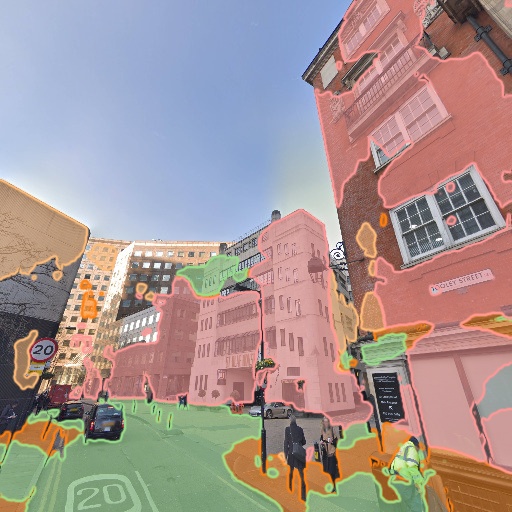}%
\includegraphics[width=\figsize,frame]{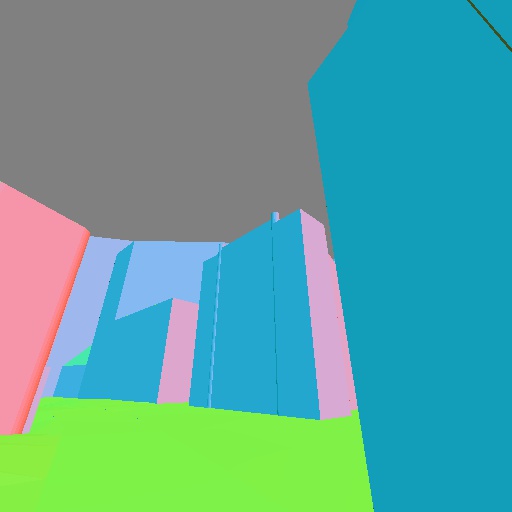}%
\includegraphics[width=\figsize,frame]{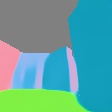}%
\includegraphics[width=\figsize,frame]{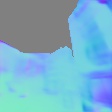}%

\includegraphics[width=\figsize,frame]{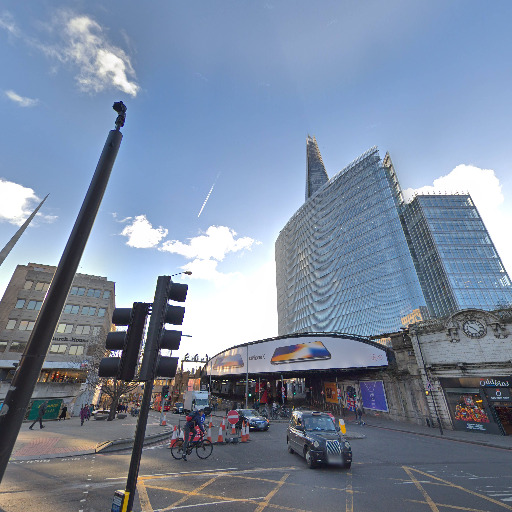}%
\includegraphics[width=\figsize,frame]{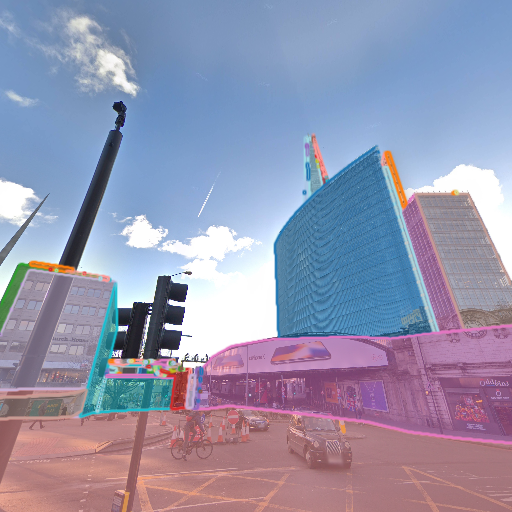}%
\includegraphics[width=\figsize,frame]{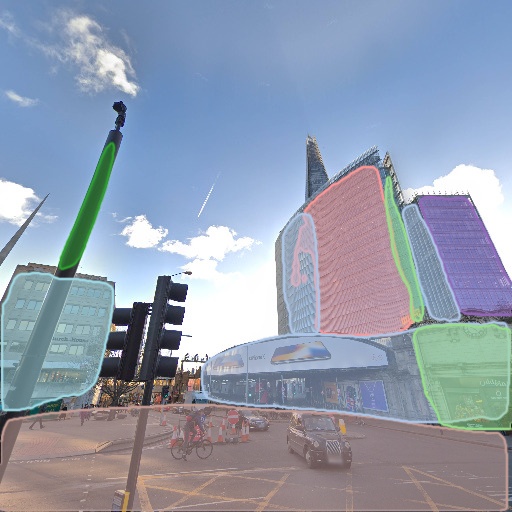}%
\includegraphics[width=\figsize,frame]{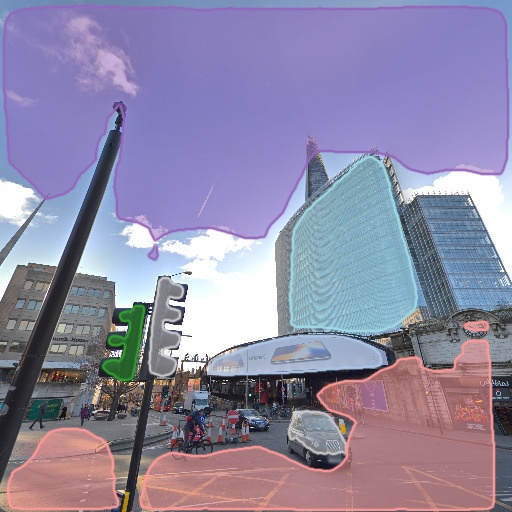}%
\includegraphics[width=\figsize,frame]{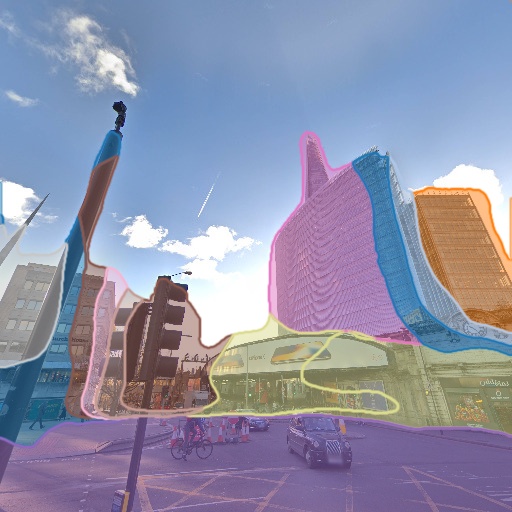}%
\includegraphics[width=\figsize,frame]{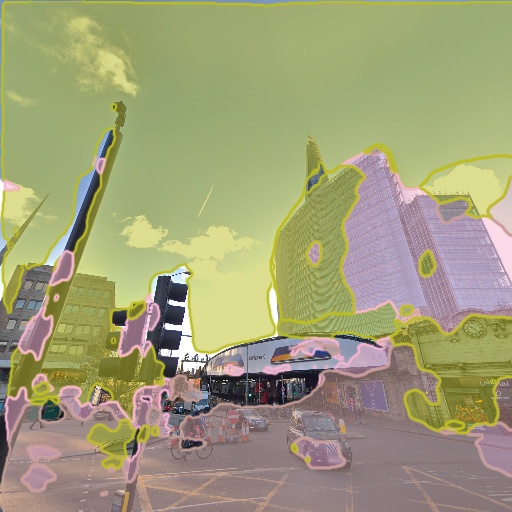}%
\includegraphics[width=\figsize,frame]{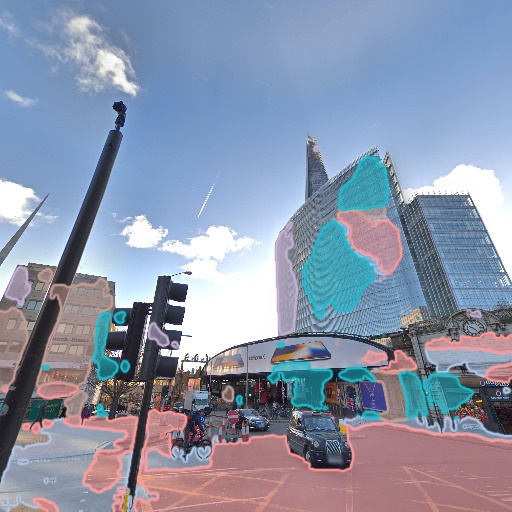}%
\includegraphics[width=\figsize,frame]{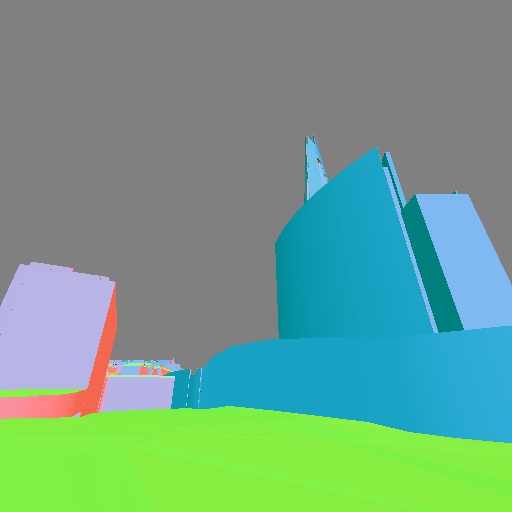}%
\includegraphics[width=\figsize,frame]{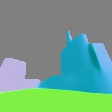}%
\includegraphics[width=\figsize,frame]{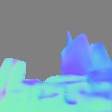}%

\includegraphics[width=\figsize,frame]{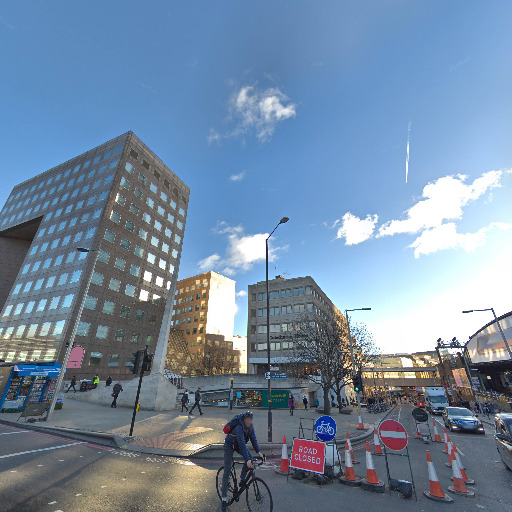}%
\includegraphics[width=\figsize,frame]{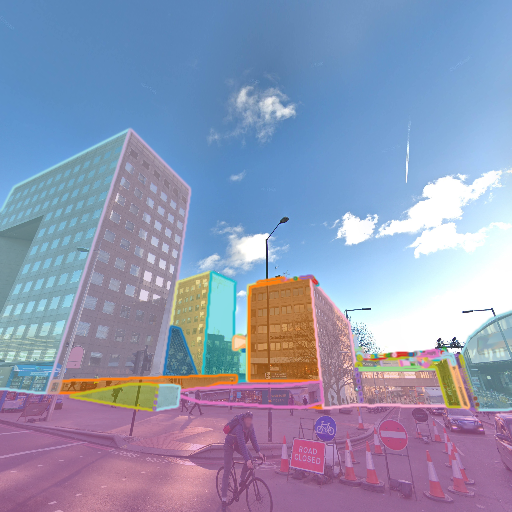}%
\includegraphics[width=\figsize,frame]{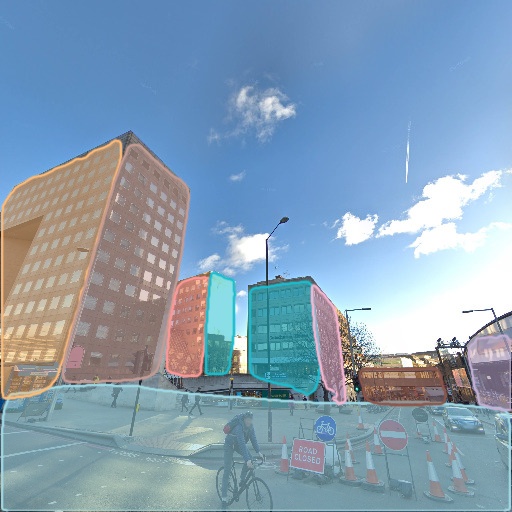}%
\includegraphics[width=\figsize,frame]{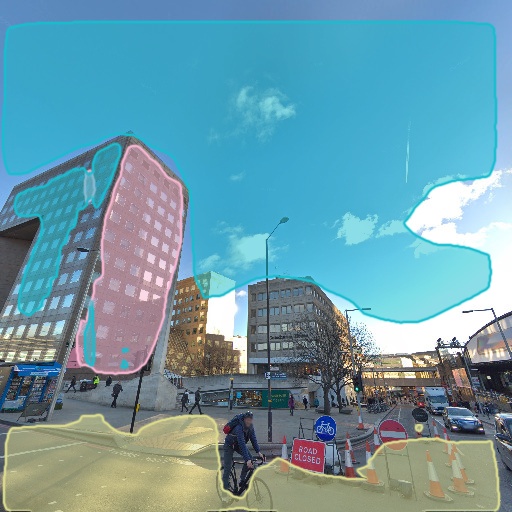}%
\includegraphics[width=\figsize,frame]{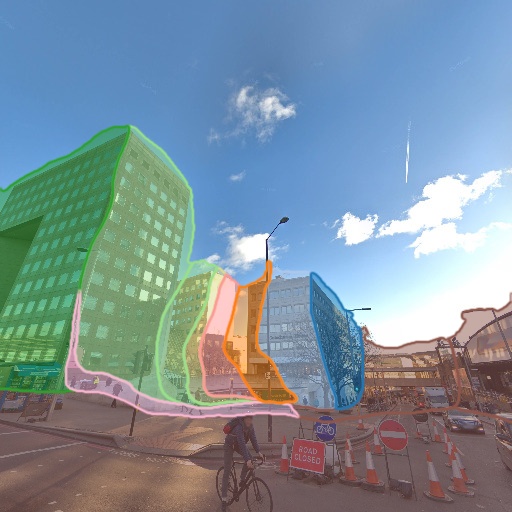}%
\includegraphics[width=\figsize,frame]{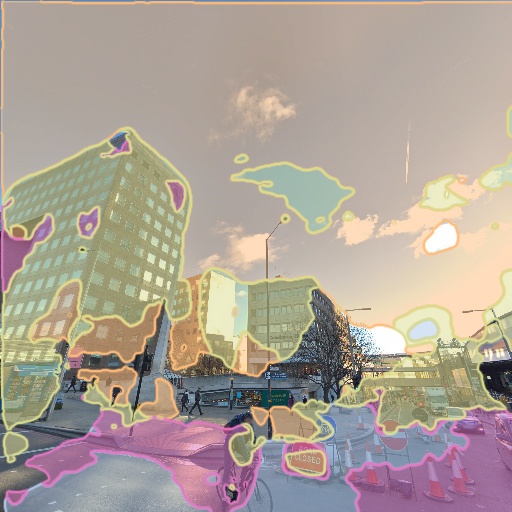}%
\includegraphics[width=\figsize,frame]{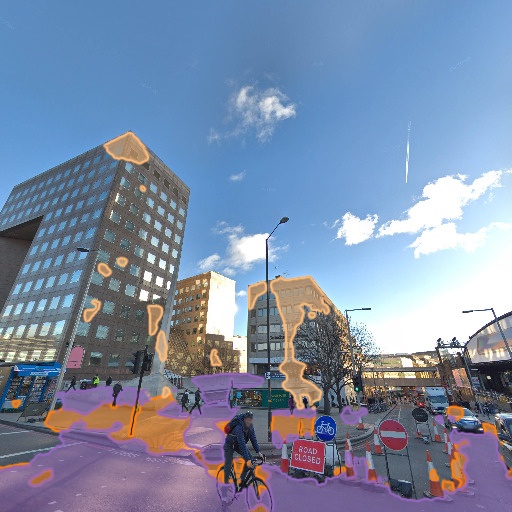}%
\includegraphics[width=\figsize,frame]{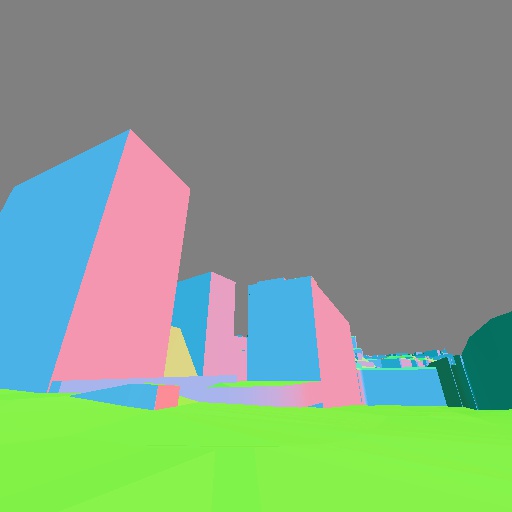}%
\includegraphics[width=\figsize,frame]{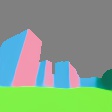}%
\includegraphics[width=\figsize,frame]{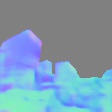}%

\vspace{2mm}%
\fontsize{7pt}{7pt}\selectfont%
\begin{minipage}{\figsize}\centering(a) RGB image\\of HoliCity\\(input)\end{minipage}%
\begin{minipage}{\figsize}\centering(b) Ground truth\\of HoliCity\\(segment)\end{minipage}%
\begin{minipage}{\figsize}\centering(c) \cite{he2017mask} trained\\on HoliCity (segment) \end{minipage}%
\begin{minipage}{\figsize}\centering(d) \cite{he2017mask} trained\\on ScanNet (segment) \end{minipage}%
\begin{minipage}{\figsize}\centering(e) \cite{yu2019single} trained\\on HoliCity (segment) \end{minipage}%
\begin{minipage}{\figsize}\centering(f) \cite{yu2019single} trained\\on ScanNet (segment)\end{minipage}%
\begin{minipage}{\figsize}\centering(g) \cite{yang2018recovering} trained\\on SYNTHIA (segment)\end{minipage}%
\begin{minipage}{\figsize}\centering(h) Ground truth\\of HoliCity\\(normal) \end{minipage}%
\begin{minipage}{\figsize}\centering(i) \cite{ronneberger2015u} trained\\on HoliCity (normal) \end{minipage}%
\begin{minipage}{\figsize}\centering(j) \cite{ronneberger2015u} trained\\on ScanNet (normal)\end{minipage}%

  \caption{Qualitative results of models evaluated on HoliCity.  We test models of MaskRCNN \cite{he2017mask}, Associative Embedding \cite{yu2019single}, PlaneRecover \cite{yang2018recovering}, and UNet \cite{ronneberger2015u} that are trained on HoliCity, ScanNet \cite{dai2017scannet}, and SYNTHIA \cite{ros2016synthia} on HoliCity.}
\label{fig:results:holicity}
\vspace{-0.1in}
\end{figure*}

\begin{figure*}[t]

\centering
\def\figsize{0.125\linewidth}

\setlength{\lineskip}{0.0ex}
\includegraphics[width=\figsize,height=\figsize,frame]{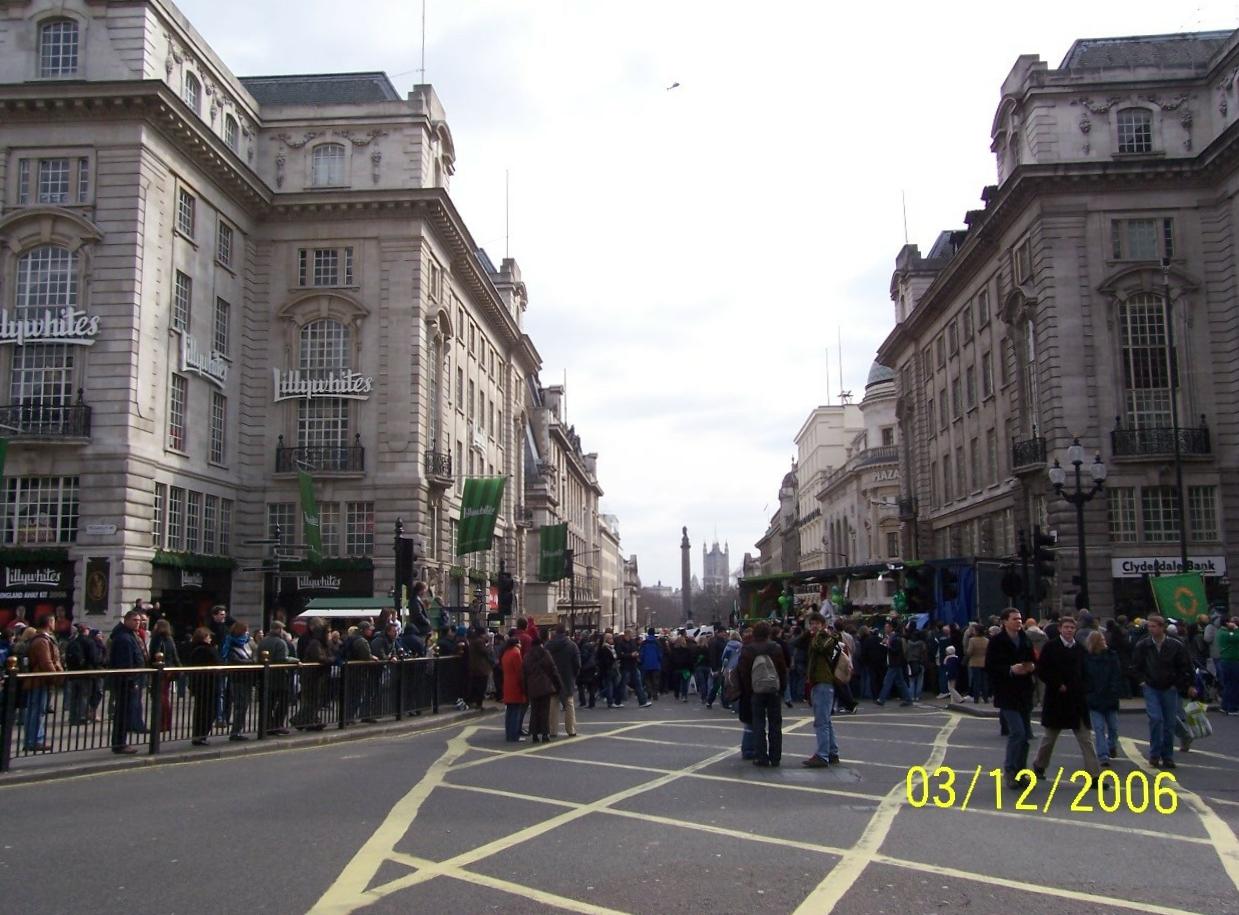}%
\includegraphics[width=\figsize,frame]{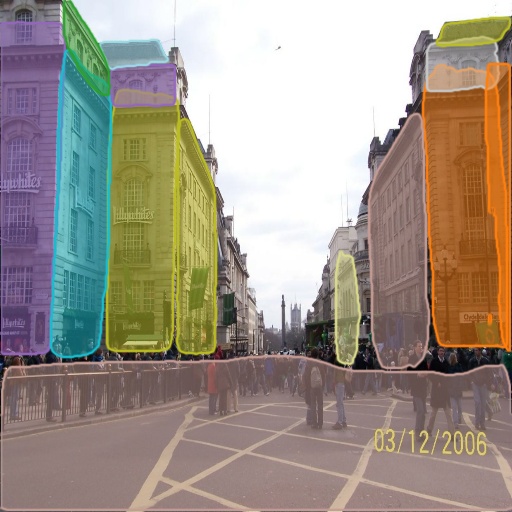}%
\includegraphics[width=\figsize,frame]{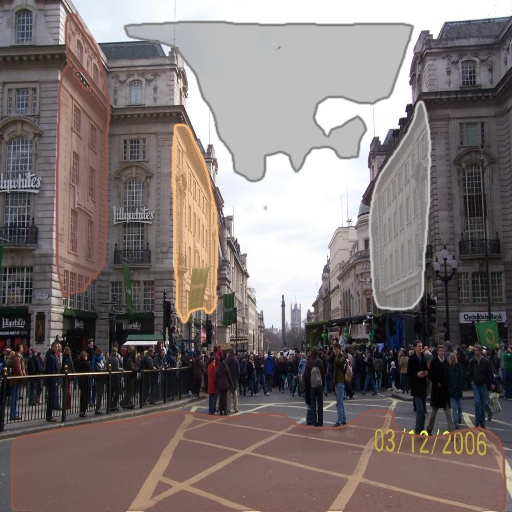}%
\includegraphics[width=\figsize,frame]{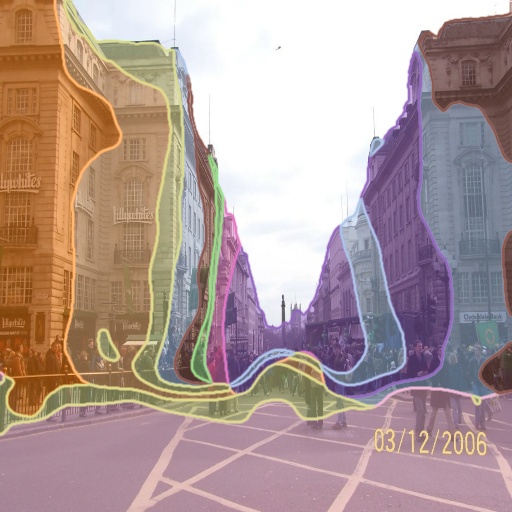}%
\includegraphics[width=\figsize,frame]{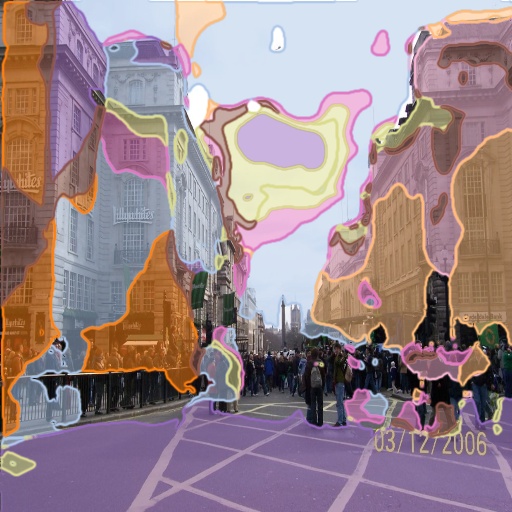}%
\includegraphics[width=\figsize,frame]{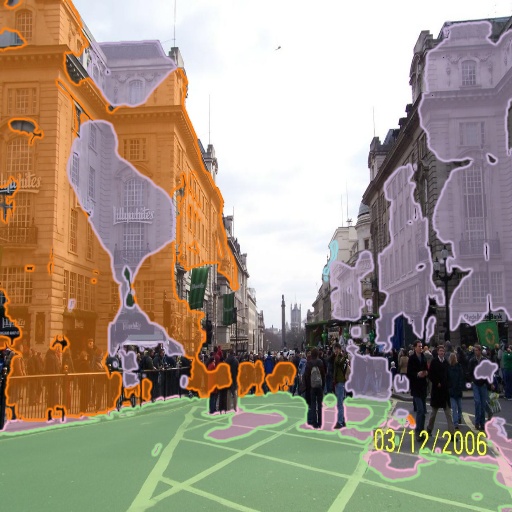}%
\includegraphics[width=\figsize,frame]{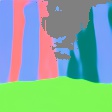}%
\includegraphics[width=\figsize,frame]{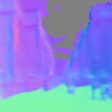}%

\includegraphics[width=\figsize,height=\figsize,frame]{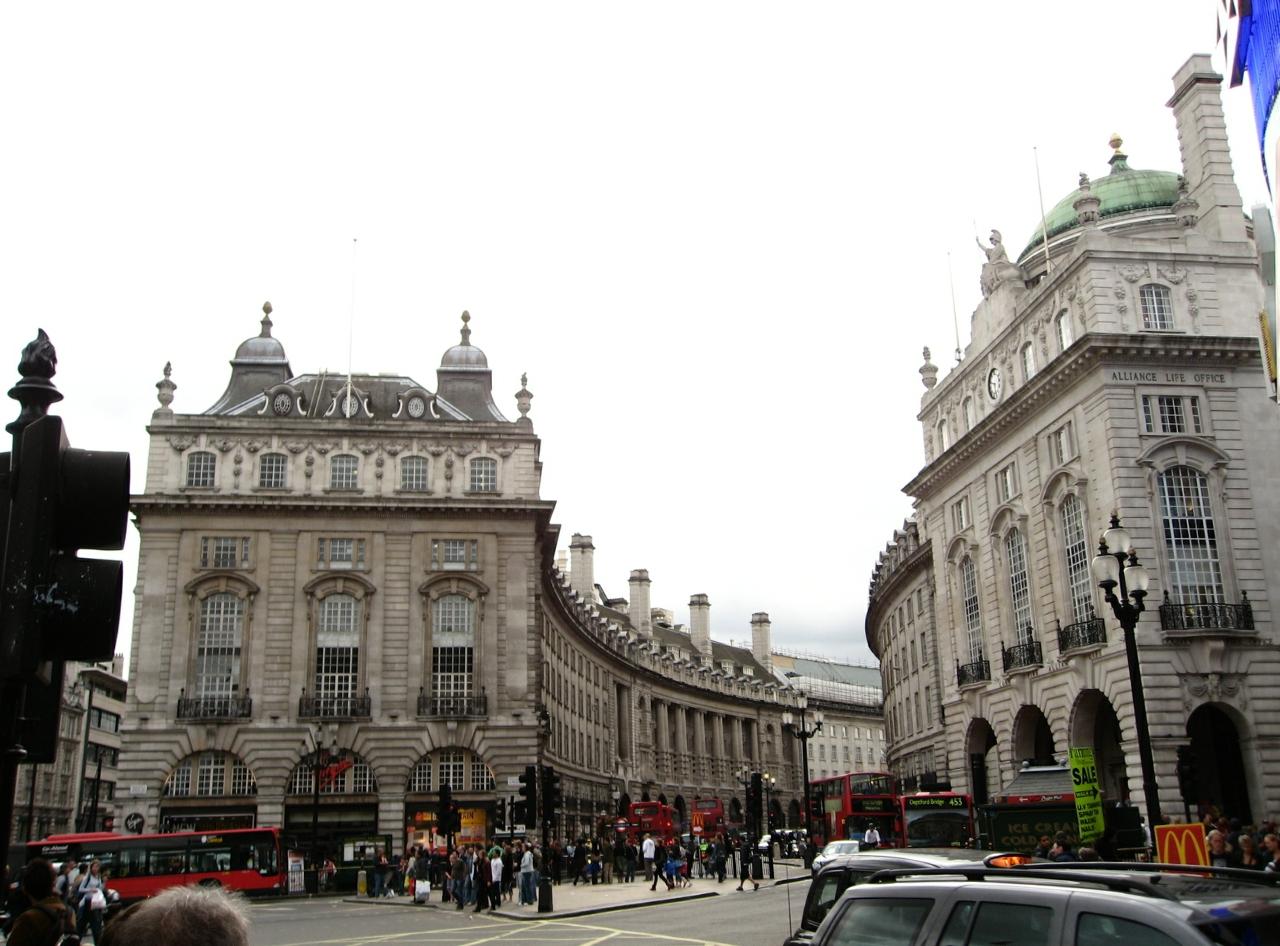}%
\includegraphics[width=\figsize,frame]{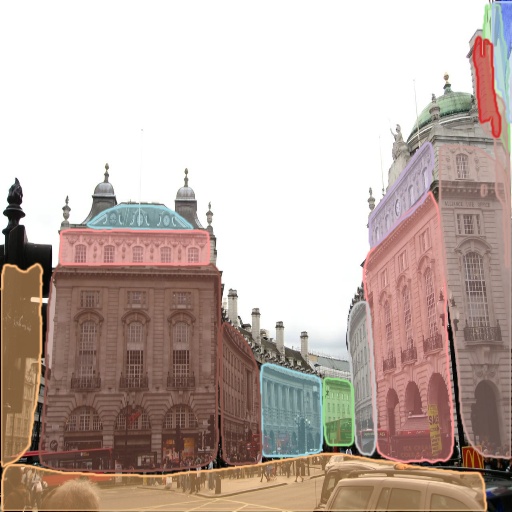}%
\includegraphics[width=\figsize,frame]{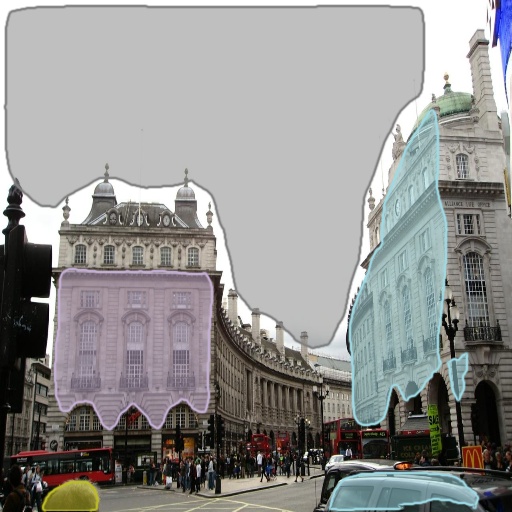}%
\includegraphics[width=\figsize,frame]{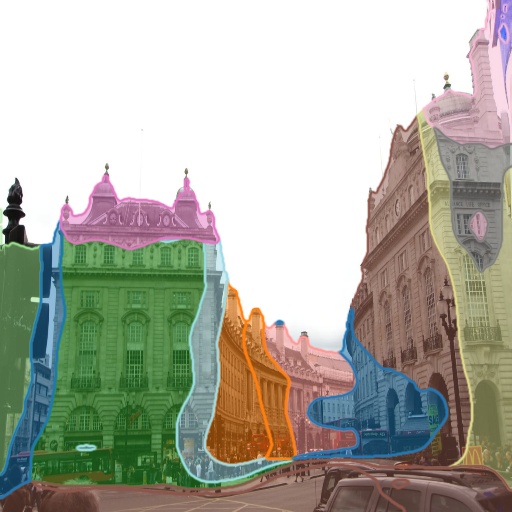}%
\includegraphics[width=\figsize,frame]{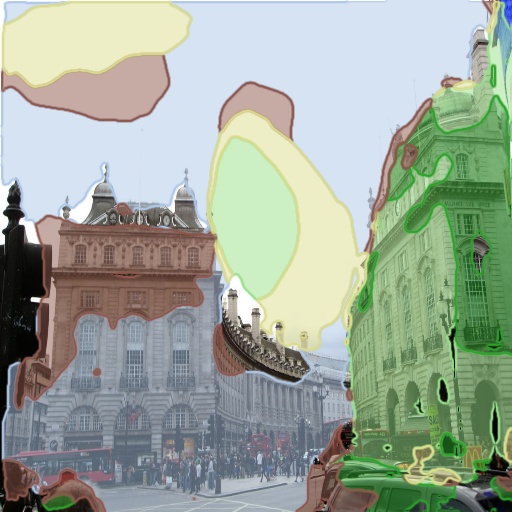}%
\includegraphics[width=\figsize,frame]{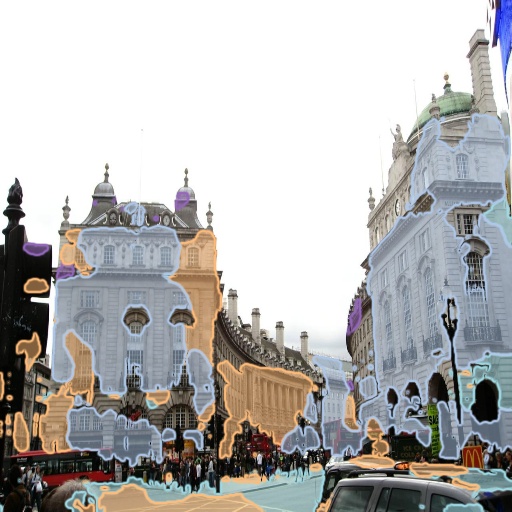}%
\includegraphics[width=\figsize,frame]{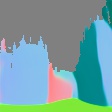}%
\includegraphics[width=\figsize,frame]{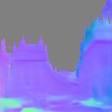}%

\vspace{2mm}
\scriptsize

\begin{minipage}{\figsize} \centering (a) RGB image \\ of MegaDepth\\  (input) \end{minipage}%
\begin{minipage}{\figsize} \centering (b) \cite{he2017mask} trained \\ on HoliCity \\ (segment) \end{minipage}%
\begin{minipage}{\figsize} \centering (c) \cite{he2017mask} trained \\ on ScanNet \\ (segment) \end{minipage}%
\begin{minipage}{\figsize} \centering (d) \cite{yu2019single} trained \\ on HoliCity \\ (normal) \end{minipage}%
\begin{minipage}{\figsize} \centering (e) \cite{yu2019single} trained \\ on ScanNet \\ (normal) \end{minipage}%
\begin{minipage}{\figsize} \centering (f) \cite{yang2018recovering} trained \\ on SYNTHIA \\ (segment) \end{minipage}%
\begin{minipage}{\figsize} \centering (g) \cite{ronneberger2015u} trained \\ on \OURS{} \\ (normal) \end{minipage}%
\begin{minipage}{\figsize} \centering (h) \cite{ronneberger2015u} trained \\ on ScanNet \\ (normal) \end{minipage}%

\caption{
  Qualitative results of models evaluated on images from the MegaDepth dataset \cite{zhengqi2018megadepth}.
  We test models of MaskRCNN \cite{he2017mask}, Associative Embedding \cite{yu2019single} and UNet \cite{ronneberger2015u} trained on \OURS{}, ScanNet, and SYNTHIA.
  Models are \emph{NOT} fine-tuned on MegaDepth.
}
\label{fig:results:MegaDepth}
\vspace{-0.1in}
\end{figure*}

\begin{figure}
\def\figsize{0.250\linewidth}
\centering
\setlength{\lineskip}{0.0ex}

\includegraphics[width=\figsize,height=\figsize,frame]{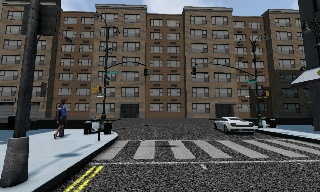}%
\includegraphics[width=\figsize,height=\figsize,frame]{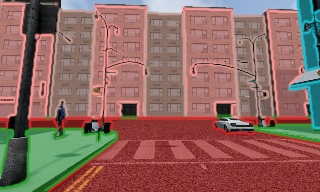}%
\includegraphics[width=\figsize,height=\figsize,frame]{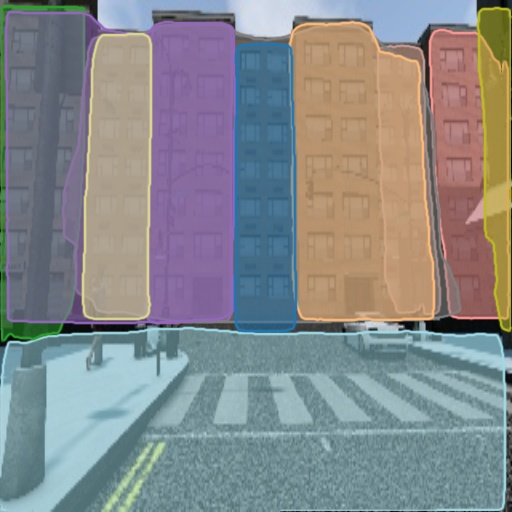}%
\includegraphics[width=\figsize,height=\figsize,frame]{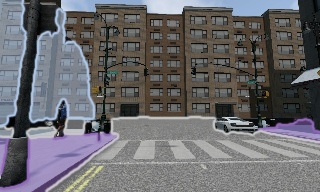}%

\includegraphics[width=\figsize,height=\figsize,frame]{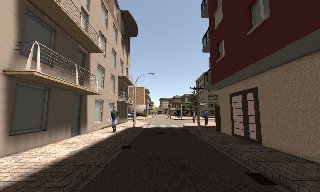}%
\includegraphics[width=\figsize,height=\figsize,frame]{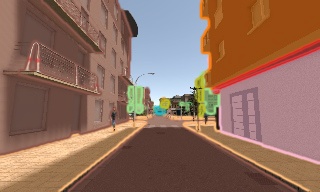}%
\includegraphics[width=\figsize,height=\figsize,frame]{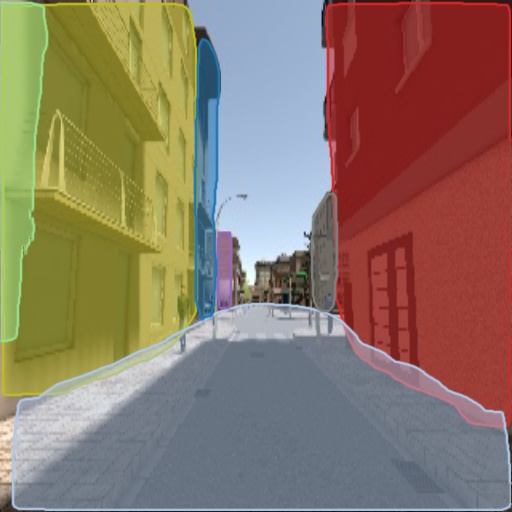}%
\includegraphics[width=\figsize,height=\figsize,frame]{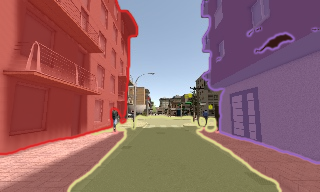}%

\vspace{2mm}
\scriptsize

\begin{minipage}{\figsize} \centering (a) RGB image \\ of SYNTHIA \\ (input) \end{minipage}%
\begin{minipage}{\figsize} \centering (b) Ground truth of SYNTHIA \\ (segment) \end{minipage}%
\begin{minipage}{\figsize} \centering (c) \cite{he2017mask} trained \\on \OURS{} (segment) \end{minipage}%
\begin{minipage}{\figsize} \centering (d) \cite{yang2018recovering} trained \\on SYNTHIA (segment) \end{minipage}%

\caption{
  Qualitative results of models evaluated on SYNTHIA.
  We test MaskRCNN and PlaneRecover \cite{yang2018recovering} trained on HoliCity and SYNTHIA.
  Models trained on HoliCity (c) are \emph{NOT} fine-tuned.
}
\label{fig:results:SYNTHIA}
\end{figure}

\section{Experiments}
In this section, we will justify the necessity of data platforms based on CAD models, e.g.,  HoliCity, for 3D vision research.
We conduct experiments on the tasks of \emph{surface segmentation} (high-level representation) and \emph{surface normal estimation} (low-level representation) to demonstrate the use of HoliCity and study of its generalizability from and to other datasets.
The reason we choose surface segmentation and normal estimation is because previously researchers hardly test their methods in outdoor environments for these tasks.
For example, existing works on surface normal estimation \cite{eigen2015predicting,bansal2016marr,wang2015designing,huang2019framenet} only demonstrate their results on indoor scenes.
We hypothesize that this is because the quasi-dense and noisy points clouds from outdoor datasets cannot reliably provide the direction of the surface normal.

For surface segmentation, algorithms take RGB images as input and predict regions that are considered as a continuous smooth surface, as shown in the second row of \Cref{fig:align}.
It can be viewed as generalized plane detection \cite{liu2018planenet}, in which curved surfaces are also included in addition to flat planes.
We note that the surface segmentation does not exclude small occluding objects such as pedestrians and poles, as we care more about the underlying 3D geometry elements rather than the 2D segmentation.
Surface segmentation is useful for 3D vision applications in AR/VR such as object placement.
Prior to HoliCity, most plane detection methods deal with indoor datasets \cite{liu2018planenet,liu2019planercnn,yu2019single} or synthetic urban scenes \cite{yang2018recovering}, probably because it is relatively hard to extract high-quality planes from noisy point clouds.  

The uses of HoliCity are not limited to the aforementioned tasks.  With CAD models, researchers have the freedom to process and convert our data into a wide range of representations and extract the structures they need.  In the supplementary material, we demonstrate other potential applications such as visual relocalization, depth estimation, and vanishing point detection with HoliCity.

\subsection{Data Processing}
\paragraph{Rendering.}
As most algorithms only support perspective images as input, we provide the perspective renderings for all the viewpoints.
For each panorama, we sample 8 views with evenly-spaced (45 degrees apart) yaw angles and randomly sampled pitch angles between 0 and 45 degrees.
We use the camera with a 90-degree field of view and render the images with resolution $512 \times 512$.
We render depth maps, normal maps, and semantic segmentation (\Cref{fig:teaser:renderings,fig:align}) from the CAD model with the same specifications using OpenGL.

\paragraph{Quality Assurance.}
Although we have made our best effort to align CAD models and images, there are still some mismatches, mostly due to construction over time and errors in the CAD models.
Therefore, we maintain a list of scenes where there is no obvious glitch visible to a human eye for assessing and assuring the quality of HoliCity.
This list contains 38705 perspective renderings, which is 77\% of all images.
Such a result indicates that our pipeline can produce a high-quality dataset, considering the rule is stringent.

\paragraph{Splitting.}
We provide two different splits of viewpoints as training, validation, and testing sets.
\begin{enumerate}
    \item data are split randomly for tasks such as relocalization;
    \item data are split according to $x$ and $y$ coordinates so that there is no spatial overlap between each set. We use it to study the generalizability of tasks such as normal estimation and surface segmentation.
\end{enumerate}

\paragraph{Surface Segmentation.}
One advantage of HoliCity over traditional LiDAR-based outdoor datasets such as KITTI \cite{geiger2013vision} and RobotCar \cite{maddern20171} is that the CAD model from HoliCity could provide a structured and accurate representation of surfaces, which makes extracting high-level representations more reliable.
Here, we briefly describe our algorithm for extracting the surface segmentation from HoliCity. The sampled results are shown in the second row of \Cref{fig:align}.

The CAD model in our dataset is represented as a set of polygons of surfaces.
We do a breadth-first-search (BFS) to compute the surface segment of each polygon.
For each nearby polygon visited during BFS, we add it into the current segments if the (approximated) curvature at the intersection line between the adjacent polygons is less than a threshold.
This threshold controls the minimal curvature required for splitting a surface segment.
Because the provided CAD model is not a perfect manifold, we treat two polygons as neighbors if there exists a vertex on each of them whose distance is smaller than a threshold distance.
This distance also controls the granularity of the resulting segments.
Increasing its value removes small segments.

\subsection{Settings and Baselines}
Although it is hard to extract high-quality surface segments and normal maps from traditional outdoor datasets, it is still possible to train models on an indoor or synthetic outdoor dataset and then apply them to a real-world outdoor dataset.
Therefore, we design experiments to evaluate its feasibility and justify the necessity of HoliCity. We also test how well the model trained on HoliCity can generalize to other street-view datasets such as MegaDepth \cite{zhengqi2018megadepth}.

\paragraph{Datasets.}
We use HoliCity (ours), ScanNet \cite{dai2017scannet} (indoor), SYNTHIA (synthetic outdoor) as the training datasets.  We evaluate the trained models on images from HoliCity, MegaDepth \cite{zhengqi2018megadepth}, and SYNTHIA.  We perform both qualitative and quantitative comparisons on HoliCity and SYNTHIA, while we only perform the qualitative comparison on street-view images of MegaDepth because the ground truth surface segmentation and surface normal are not provided.

\paragraph{Surface Segmentation.}
We include three baseline methods: MaskRCNN \cite{he2017mask}, Associative Embedding \cite{yu2019single}, and PlaneRecover \cite{yang2018recovering}.
MaskRCNN is the most popular method for instance segmentation.
We use the implementation from Detectron2 \cite{wu2019detectron2} and train the models by ourselves.
Associative Embedding is a method for indoor plane detection.
We use its official pre-trained model on ScanNet and retrain the Associative Embedding model on HoliCity from scratches.
PlaneRecover is an approach designed for SYNTHIA \cite{ros2016synthia}. We evaluate its official pre-trained model.

\paragraph{Normal Estimation.}
We report the performance of UNet \cite{ronneberger2015u}.  We train the models on all datasets by ourselves.  %

\subsection{Results and Discussions}
We show the qualitative results evaluated on the HoliCity dataset of multiple methods in \Cref{fig:results:holicity}, in which we trained the models of MaskRCNN \cite{he2017mask}, Associative Embedding \cite{yu2019single}, PlaneRecover \cite{yang2018recovering}, and UNet \cite{ronneberger2015u} on HoliCity (ours) ScanNet \cite{dai2017scannet} (indoor dataset), and SYNTHIA \cite{ros2016synthia} (synthetic outdoor dataset) on the task of surface segmentation and normal estimation.
We find that for both tasks methods trained on ScanNet and SYNTHIA do not generalize well to HoliCity, which is probably due to the domain gap between training sets and testing sets.
This can also be verified by the quantitative metrics in \Cref{tab:results:HoliCity}.
We can see that the methods trained on indoor or synthetic outdoor datasets perform much worse on real-world outdoor scenes than the methods trained on HoliCity.
We conclude that for existing methods such as MaskRCNN and Associative Embedding, a dataset such as HoliCity is necessary for the tasks of surface segmentation and normal estimation in outdoor environments.

\begin{table}
  \renewcommand{\arraystretch}{1.1}
  \centering
  \resizebox{\linewidth}{!}{%
\begin{tabular}{c|c|ccc|c}
\hline
\hline
\multirow{2}{*}{\begin{tabular}[c]{@{}c@{}}Methods\end{tabular}}                       & \multirow{2}{*}{\begin{tabular}[c]{@{}c@{}}Training\\ Datasets\end{tabular}} & \multicolumn{3}{c|}{Surface Segmentation} & Normal Est. \\ \cline{3-6} 
                                                                                                   &                                                                              & AP${}_{50}$    & AP${}_{75}$    & mAP     & Mean Error  \\ \hline
\multirow{2}{*}{MaskRCNN \cite{he2017mask}}                                                        & HoliCity                                                                     & 42.0           & 19.8           & 21.9    &             \\
                                                                                                   & ScanNet                                                                      & 5.0            & 0.6           & 1.7    &             \\ \hline
\multirow{2}{*}{\begin{tabular}[c]{@{}c@{}}Associative\\ Embedding \cite{yu2019single}\end{tabular}} & HoliCity                                                                     & 20.2           & 8.5           & 9.9    &             \\
                                                                                                   & ScanNet                                                                      & 3.3           & 0.6            & 1.1      &             \\ \hline
  \multirow{2}{*}{UNet \cite{ronneberger2015u}}                                                      & HoliCity                                                                     &                &                &         & 22.6${}^{\circ}$        \\
                                                                                                   & ScanNet                                                                      &                &                &         & 46.3${}^{\circ}$         \\
\hline
\hline
\end{tabular}%
  }
  \caption{
    Results of surface segmentation and normal estimation evaluated on the validation split of HoliCity.
    Methods are trained on HoliCity (our dataset), ScanNet (indoor dataset) \cite{dai2017scannet}, and SYNTHIA \cite{ros2016synthia} (synthetic outdoor dataset) and tested on HoliCity \emph{without fine-tuning}. We report the AP metrics for surface segmentation and mean angular error for normal estimation.
  }
  \label{tab:results:HoliCity}
\end{table}

\begin{table}
\renewcommand{\arraystretch}{1.1}
\centering
\resizebox{\linewidth}{!}{%
\begin{tabular}{c|cc}
  \hline \hline
  \multirow{2}{*}{Training Datasets (Methods)}       & \multicolumn{2}{c}{Testing Datasets (AP${}_{50}$)} \\ \cline{2-3} 
                                                     & HoliCity           & SYNTHIA           \\ \hline
        HoliCity (MaskRCNN \cite{he2017mask})        & 42.0               & 36.1              \\
        SYNTHIA (PlaneRecover \cite{yang2018recovering}) & 1.90               & 40.6              \\ \hline
  \hline
  \end{tabular}%
}
\caption{Results of surface segmentation cross-trained and evaluated on the validation split of HoliCity and SYNTHIA \cite{ros2016synthia}.  We test our MaskRCNN model \cite{he2017mask} trained on HoliCity and the official PlaneRecover model trained on SYNTHIA from \cite{yang2018recovering}.  Models are \emph{not fine-tuned} on testing datasets.}
\label{tab:results:crossfile}
\end{table}

We also conduct the cross-dataset experiment on HoliCity and synthetic SYNTHIA datasets for surface segmentation.
In this experiment, we use the official plane detection model trained on SYNTHIA from \cite{yang2018recovering} and train the MaskRCNN \cite{he2017mask} model on HoliCity.
Then, we evaluate both models on HoliCity and SYNTHIA.
We show the quantitative results in \Cref{tab:results:crossfile}.
We find that the model trained on HoliCity can generalize to a synthetic outdoor dataset such as SYNTHIA well, while the model trained on SYNTHIA completely fails on HoliCity.
Such observations also apply to the qualitative results in \Cref{fig:results:holicity,fig:results:SYNTHIA}, where the HoliCity-trained model recovers most of the building surfaces in SYNTHIA despite the differences between the definitions of surface segments in HoliCity and planes in \cite{yang2018recovering}.
We hypothesize that the causes of these phenomena are due to the wider variety of scenes covered by HoliCity, compared to the scenes from SYNTHIA.

In fact, methods trained on ScanNet and SYNTHIA do not generalize well to HoliCity, nor to other outdoor datasets such as MegaDepth \cite{zhengqi2018megadepth}, as shown in \Cref{fig:results:MegaDepth}.
In comparison, methods trained on HoliCity produce much better surface segmentation and normal maps, which shows HoliCity's potential generalizability to general outdoor imagery. Finally, we summarize our observations as follows:
\begin{enumerate}
  \item previous research of plane detection and normal estimation hardly experiments on outdoor datasets;
  \item HoliCity can provide both high-quality holistic structures (e.g., surface segments) and low-level representations (e.g., normal maps) of urban environments;
  \item models trained on indoor or synthetic outdoor datasets cannot generalize well to real-world outdoor datasets;
  \item models trained on HoliCity can generalize to both synthetic outdoor scenes and real-world street-view imagery from different datasets.
\end{enumerate}
\simplify{These observations indicate that HoliCity is an indispensable data platform of urban environments for future research of 3D vision.}

{
  \bibliographystyle{ieee_fullname}
  \bibliography{main}
}

\clearpage
\appendix
\section{Supplementary Material}

\subsection{Random Sampled Visualization of HoliCity}

\begin{figure*}[p]
\centering
\setlength{\lineskip}{0.0ex}
\includegraphics[width=.12\linewidth]{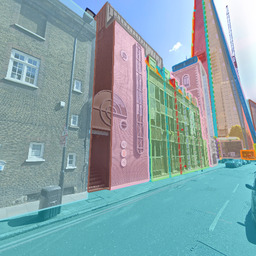}\includegraphics[width=.12\linewidth]{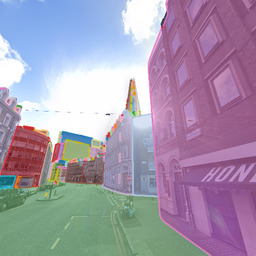}\includegraphics[width=.12\linewidth]{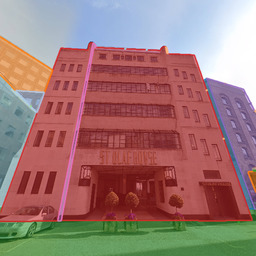}\includegraphics[width=.12\linewidth]{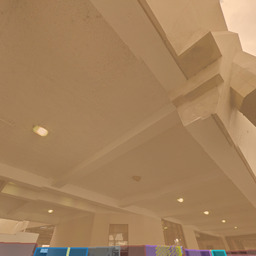}\includegraphics[width=.12\linewidth]{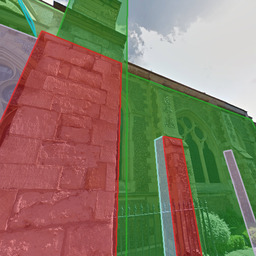}\includegraphics[width=.12\linewidth]{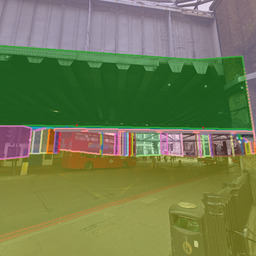}\includegraphics[width=.12\linewidth]{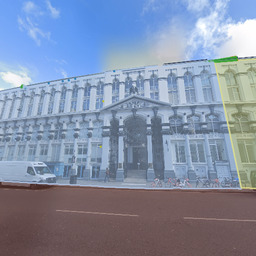}\includegraphics[width=.12\linewidth]{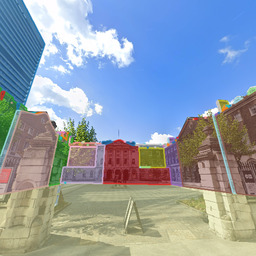}\\
\includegraphics[width=.12\linewidth]{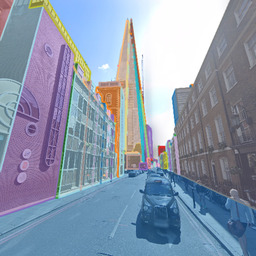}\includegraphics[width=.12\linewidth]{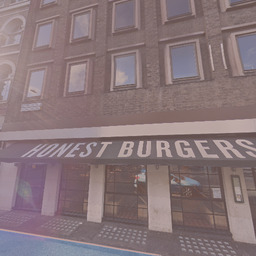}\includegraphics[width=.12\linewidth]{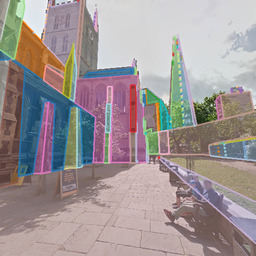}\includegraphics[width=.12\linewidth]{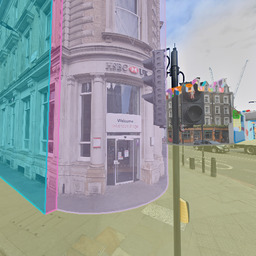}\includegraphics[width=.12\linewidth]{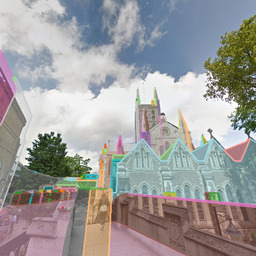}\includegraphics[width=.12\linewidth]{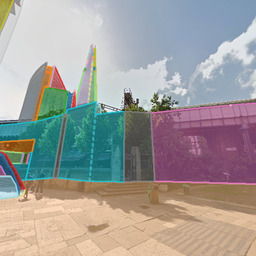}\includegraphics[width=.12\linewidth]{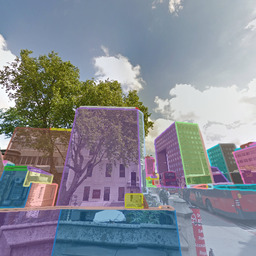}\includegraphics[width=.12\linewidth]{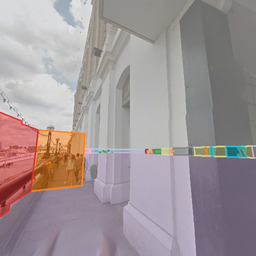}\\
\includegraphics[width=.12\linewidth]{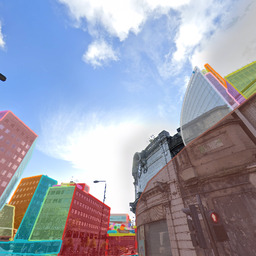}\includegraphics[width=.12\linewidth]{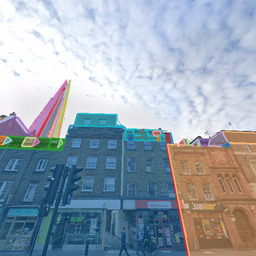}\includegraphics[width=.12\linewidth]{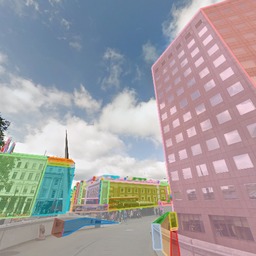}\includegraphics[width=.12\linewidth]{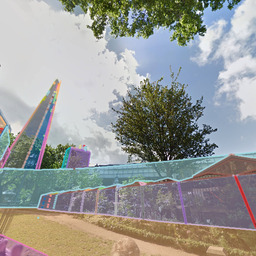}\includegraphics[width=.12\linewidth]{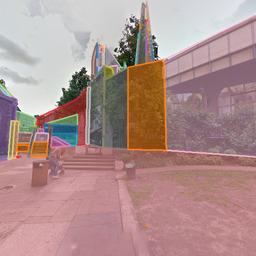}\includegraphics[width=.12\linewidth]{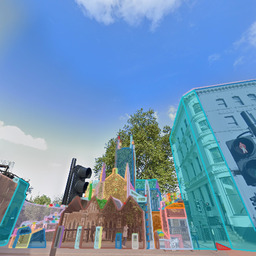}\includegraphics[width=.12\linewidth]{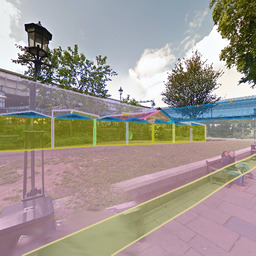}\includegraphics[width=.12\linewidth]{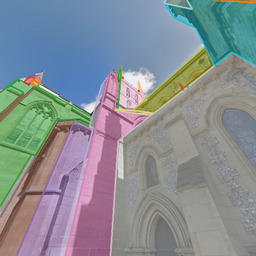}\\
\includegraphics[width=.12\linewidth]{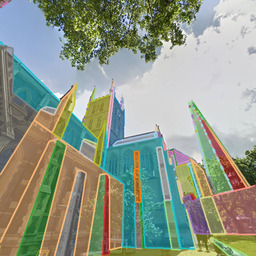}\includegraphics[width=.12\linewidth]{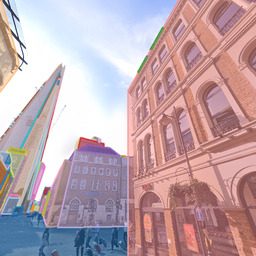}\includegraphics[width=.12\linewidth]{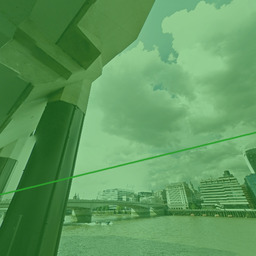}\includegraphics[width=.12\linewidth]{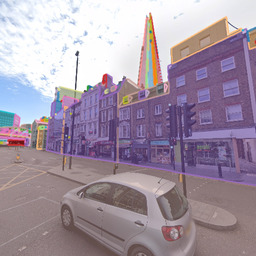}\includegraphics[width=.12\linewidth]{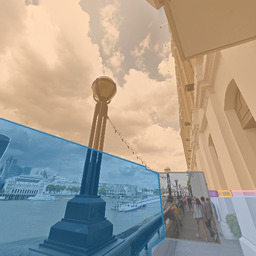}\includegraphics[width=.12\linewidth]{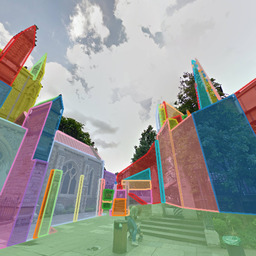}\includegraphics[width=.12\linewidth]{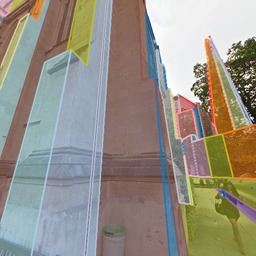}\includegraphics[width=.12\linewidth]{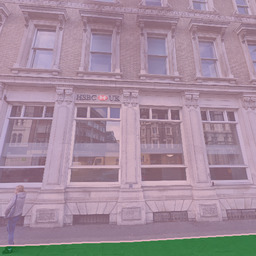}\\
\includegraphics[width=.12\linewidth]{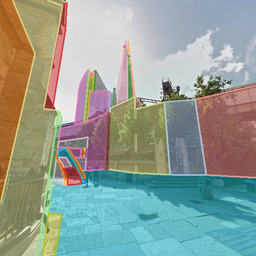}\includegraphics[width=.12\linewidth]{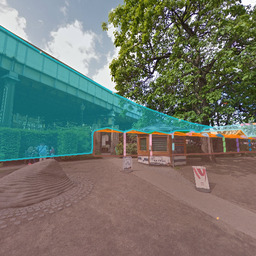}\includegraphics[width=.12\linewidth]{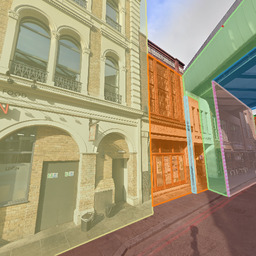}\includegraphics[width=.12\linewidth]{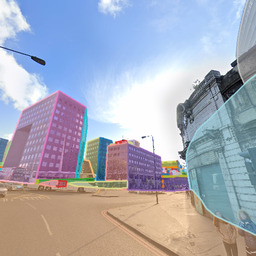}\includegraphics[width=.12\linewidth]{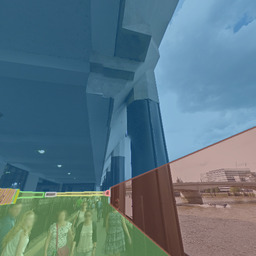}\includegraphics[width=.12\linewidth]{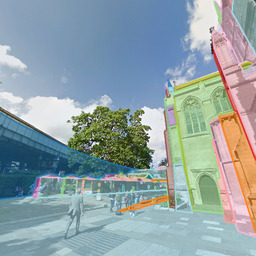}\includegraphics[width=.12\linewidth]{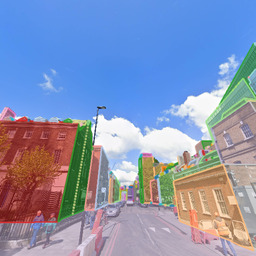}\includegraphics[width=.12\linewidth]{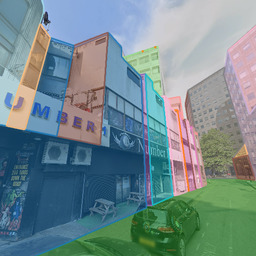}\\
\includegraphics[width=.12\linewidth]{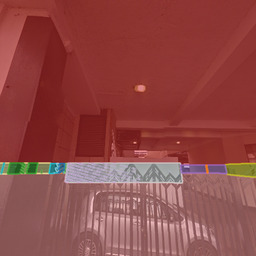}\includegraphics[width=.12\linewidth]{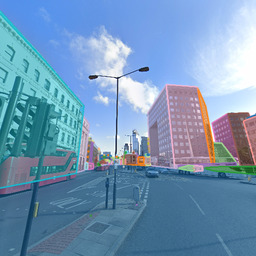}\includegraphics[width=.12\linewidth]{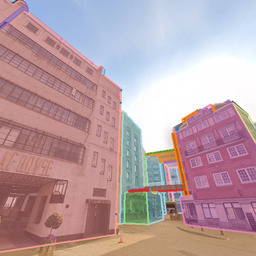}\includegraphics[width=.12\linewidth]{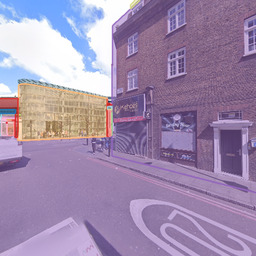}\includegraphics[width=.12\linewidth]{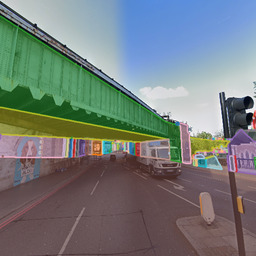}\includegraphics[width=.12\linewidth]{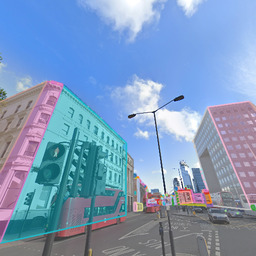}\includegraphics[width=.12\linewidth]{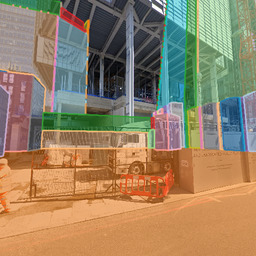}\includegraphics[width=.12\linewidth]{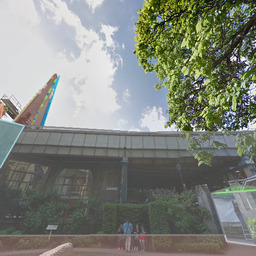}\\
\includegraphics[width=.12\linewidth]{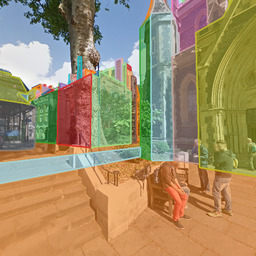}\includegraphics[width=.12\linewidth]{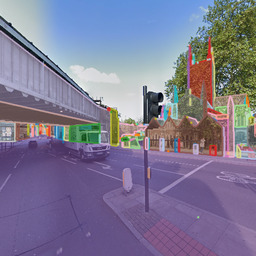}\includegraphics[width=.12\linewidth]{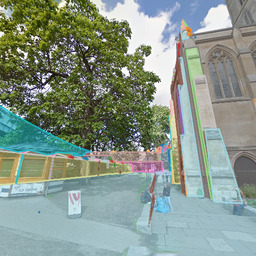}\includegraphics[width=.12\linewidth]{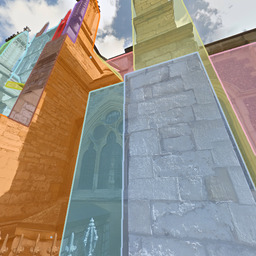}\includegraphics[width=.12\linewidth]{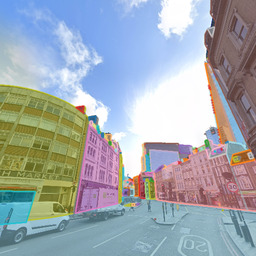}\includegraphics[width=.12\linewidth]{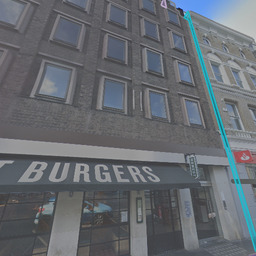}\includegraphics[width=.12\linewidth]{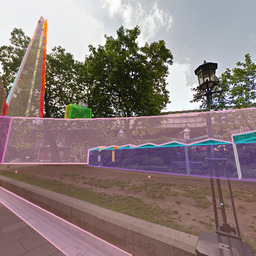}\includegraphics[width=.12\linewidth]{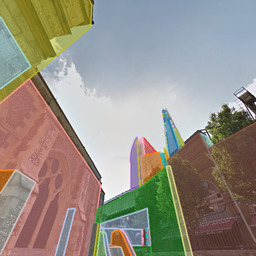}\\
\includegraphics[width=.12\linewidth]{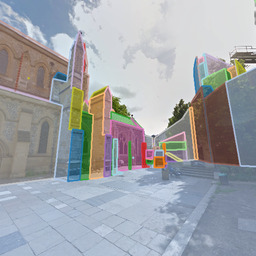}\includegraphics[width=.12\linewidth]{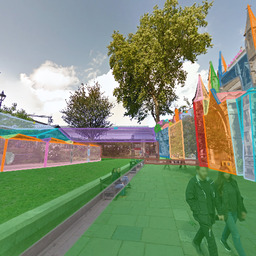}\includegraphics[width=.12\linewidth]{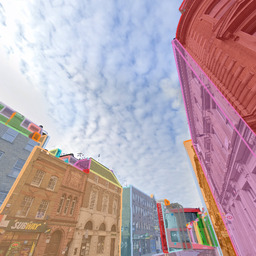}\includegraphics[width=.12\linewidth]{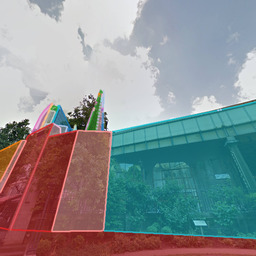}\includegraphics[width=.12\linewidth]{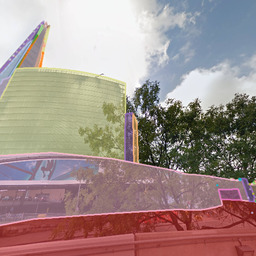}\includegraphics[width=.12\linewidth]{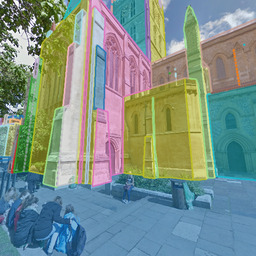}\includegraphics[width=.12\linewidth]{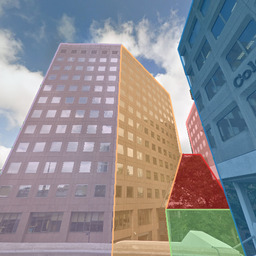}\includegraphics[width=.12\linewidth]{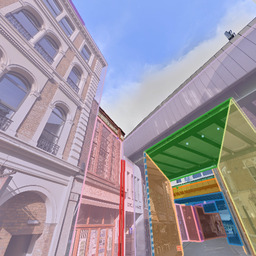}\\
\includegraphics[width=.12\linewidth]{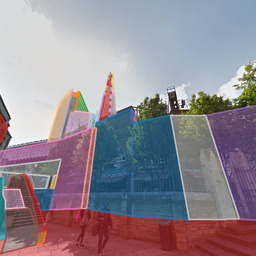}\includegraphics[width=.12\linewidth]{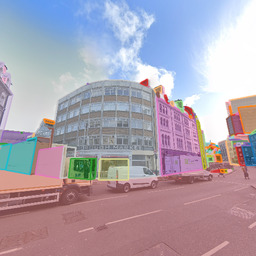}\includegraphics[width=.12\linewidth]{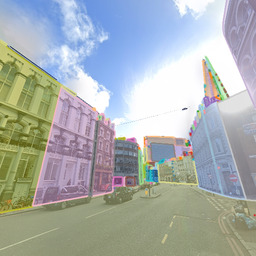}\includegraphics[width=.12\linewidth]{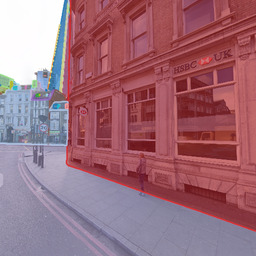}\includegraphics[width=.12\linewidth]{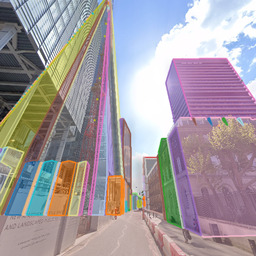}\includegraphics[width=.12\linewidth]{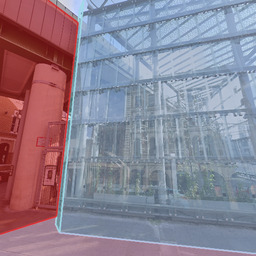}\includegraphics[width=.12\linewidth]{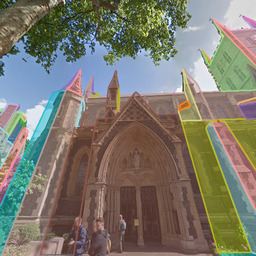}\includegraphics[width=.12\linewidth]{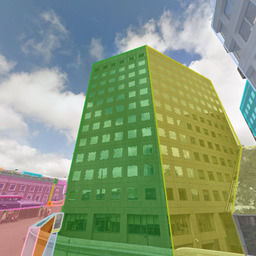}\\
\includegraphics[width=.12\linewidth]{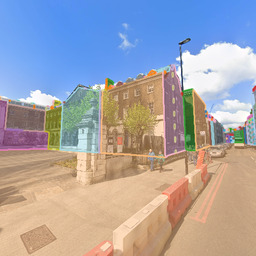}\includegraphics[width=.12\linewidth]{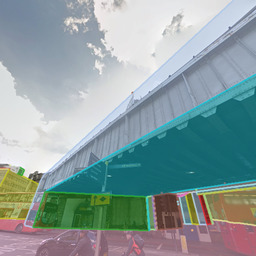}\includegraphics[width=.12\linewidth]{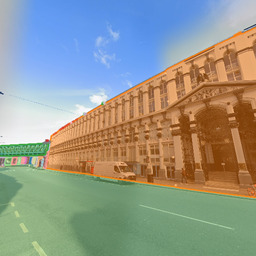}\includegraphics[width=.12\linewidth]{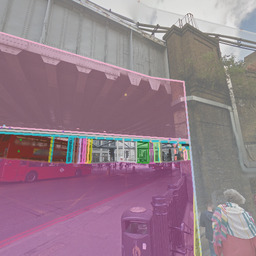}\includegraphics[width=.12\linewidth]{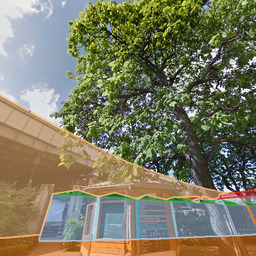}\includegraphics[width=.12\linewidth]{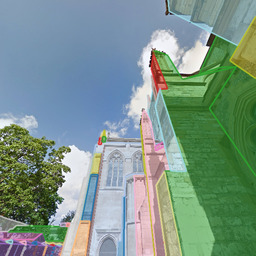}\includegraphics[width=.12\linewidth]{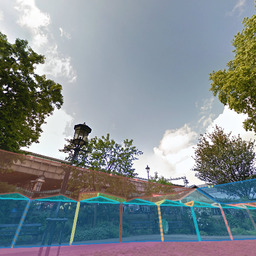}\includegraphics[width=.12\linewidth]{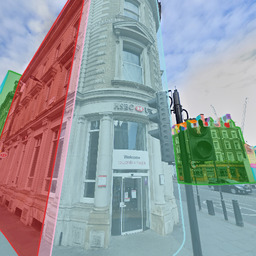}\\

\caption{\textbf{Random sampled} perspective images overlaid with surface segments from \textbf{high-resolution} CAD models.}
\label{fig:datasets:high}
\end{figure*}

\begin{figure*}[p]
\centering
\setlength{\lineskip}{0.0ex}
\includegraphics[width=.12\linewidth]{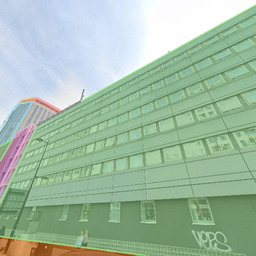}\includegraphics[width=.12\linewidth]{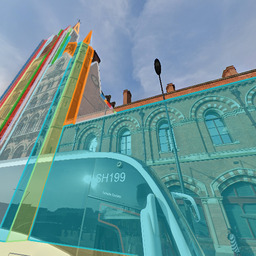}\includegraphics[width=.12\linewidth]{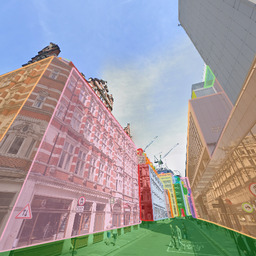}\includegraphics[width=.12\linewidth]{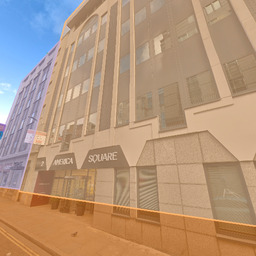}\includegraphics[width=.12\linewidth]{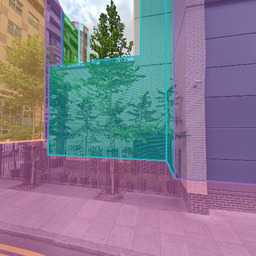}\includegraphics[width=.12\linewidth]{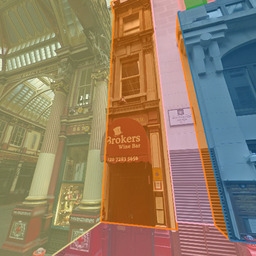}\includegraphics[width=.12\linewidth]{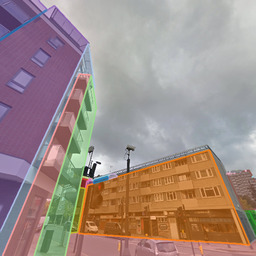}\includegraphics[width=.12\linewidth]{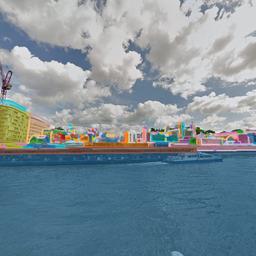}\\
\includegraphics[width=.12\linewidth]{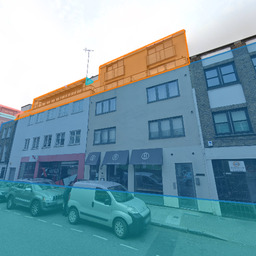}\includegraphics[width=.12\linewidth]{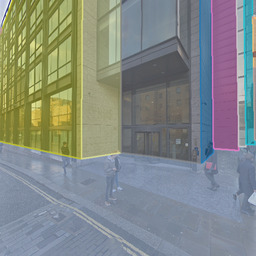}\includegraphics[width=.12\linewidth]{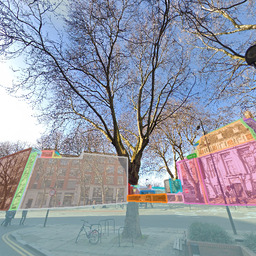}\includegraphics[width=.12\linewidth]{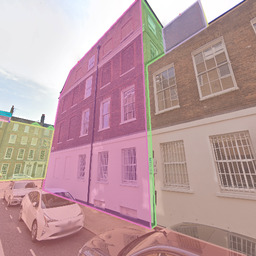}\includegraphics[width=.12\linewidth]{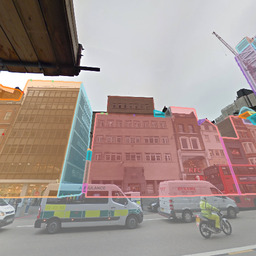}\includegraphics[width=.12\linewidth]{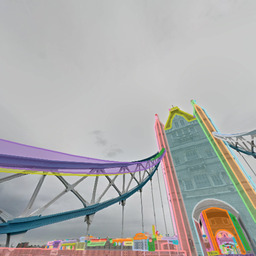}\includegraphics[width=.12\linewidth]{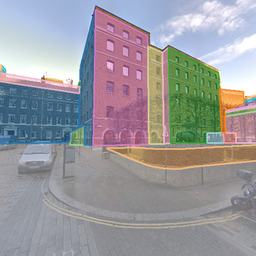}\includegraphics[width=.12\linewidth]{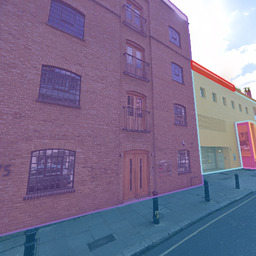}\\
\includegraphics[width=.12\linewidth]{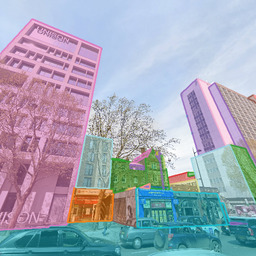}\includegraphics[width=.12\linewidth]{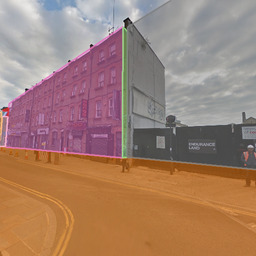}\includegraphics[width=.12\linewidth]{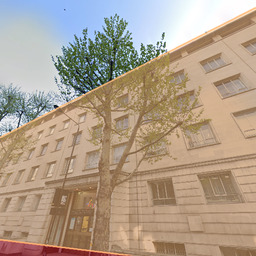}\includegraphics[width=.12\linewidth]{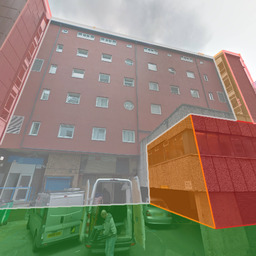}\includegraphics[width=.12\linewidth]{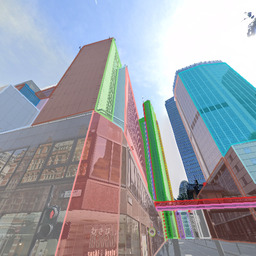}\includegraphics[width=.12\linewidth]{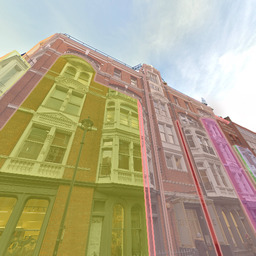}\includegraphics[width=.12\linewidth]{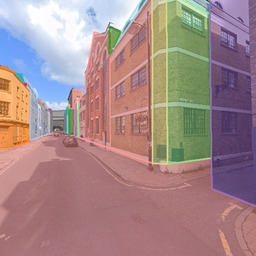}\includegraphics[width=.12\linewidth]{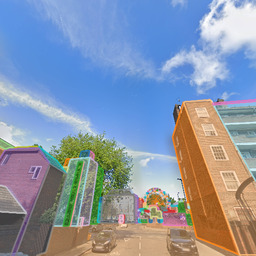}\\
\includegraphics[width=.12\linewidth]{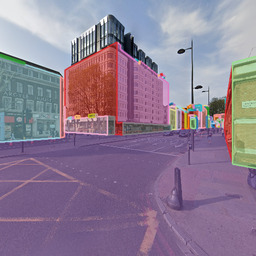}\includegraphics[width=.12\linewidth]{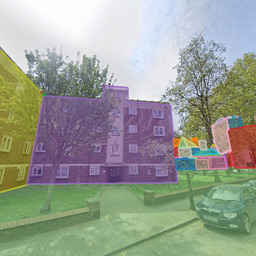}\includegraphics[width=.12\linewidth]{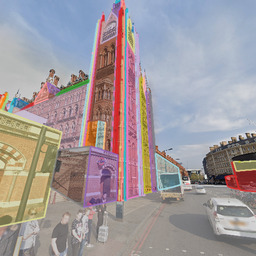}\includegraphics[width=.12\linewidth]{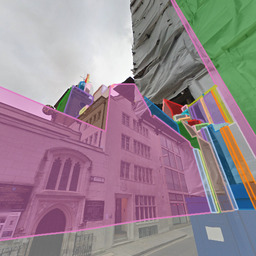}\includegraphics[width=.12\linewidth]{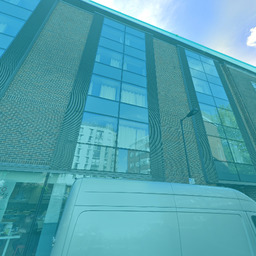}\includegraphics[width=.12\linewidth]{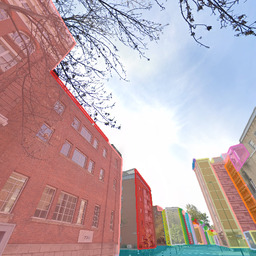}\includegraphics[width=.12\linewidth]{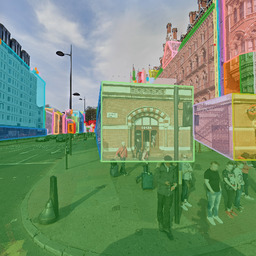}\includegraphics[width=.12\linewidth]{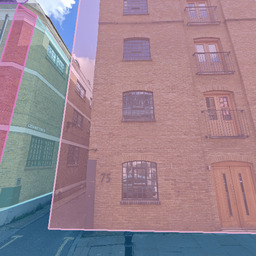}\\
\includegraphics[width=.12\linewidth]{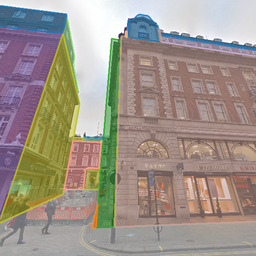}\includegraphics[width=.12\linewidth]{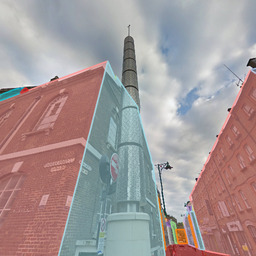}\includegraphics[width=.12\linewidth]{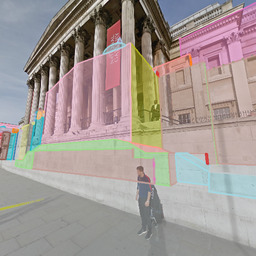}\includegraphics[width=.12\linewidth]{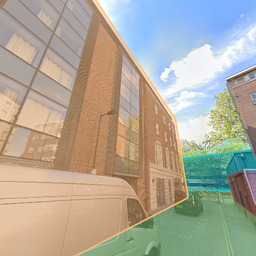}\includegraphics[width=.12\linewidth]{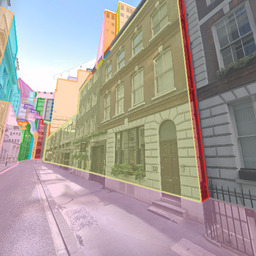}\includegraphics[width=.12\linewidth]{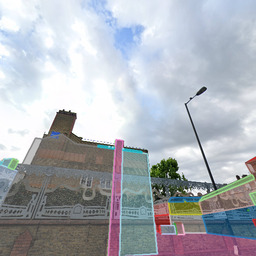}\includegraphics[width=.12\linewidth]{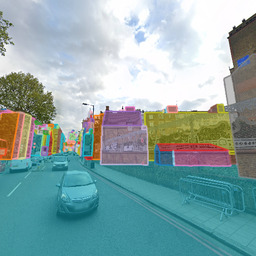}\includegraphics[width=.12\linewidth]{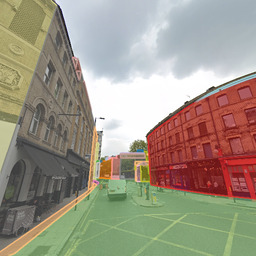}\\
\includegraphics[width=.12\linewidth]{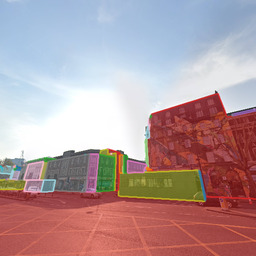}\includegraphics[width=.12\linewidth]{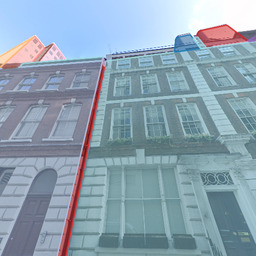}\includegraphics[width=.12\linewidth]{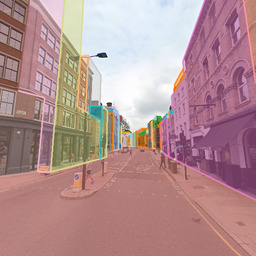}\includegraphics[width=.12\linewidth]{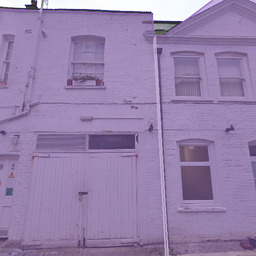}\includegraphics[width=.12\linewidth]{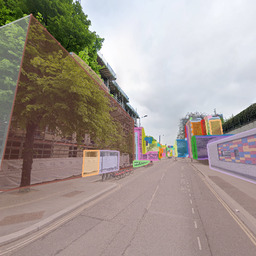}\includegraphics[width=.12\linewidth]{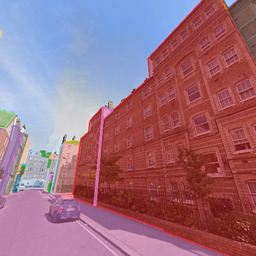}\includegraphics[width=.12\linewidth]{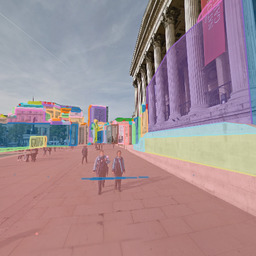}\includegraphics[width=.12\linewidth]{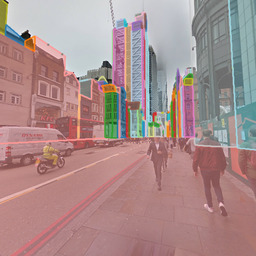}\\
\includegraphics[width=.12\linewidth]{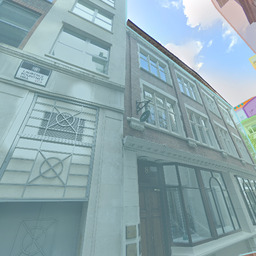}\includegraphics[width=.12\linewidth]{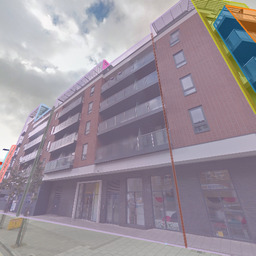}\includegraphics[width=.12\linewidth]{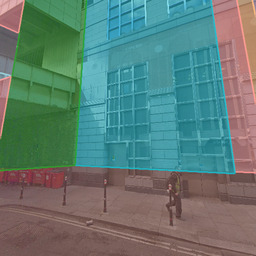}\includegraphics[width=.12\linewidth]{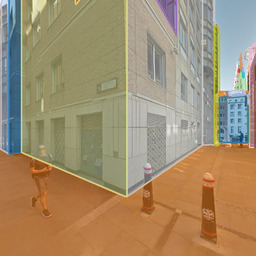}\includegraphics[width=.12\linewidth]{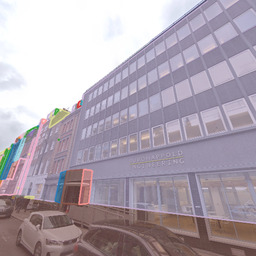}\includegraphics[width=.12\linewidth]{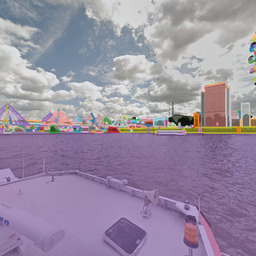}\includegraphics[width=.12\linewidth]{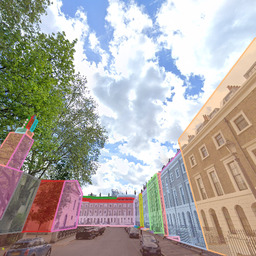}\includegraphics[width=.12\linewidth]{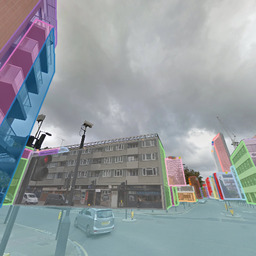}\\
\includegraphics[width=.12\linewidth]{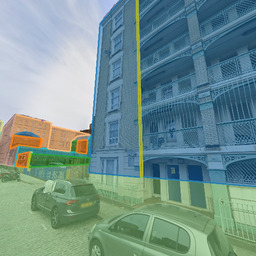}\includegraphics[width=.12\linewidth]{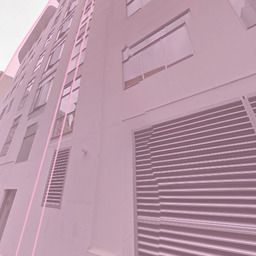}\includegraphics[width=.12\linewidth]{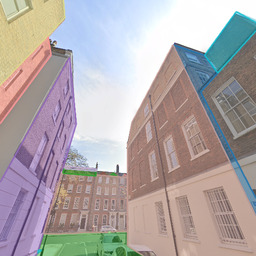}\includegraphics[width=.12\linewidth]{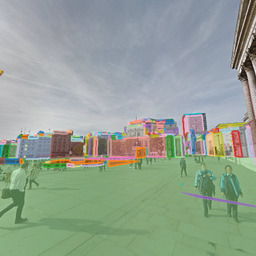}\includegraphics[width=.12\linewidth]{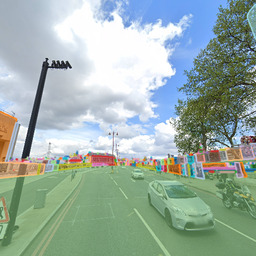}\includegraphics[width=.12\linewidth]{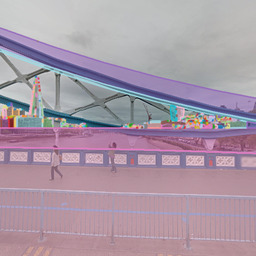}\includegraphics[width=.12\linewidth]{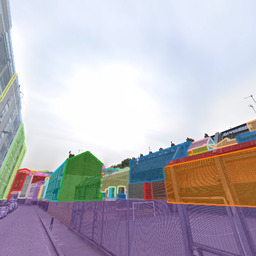}\includegraphics[width=.12\linewidth]{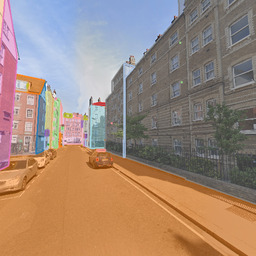}\\
\includegraphics[width=.12\linewidth]{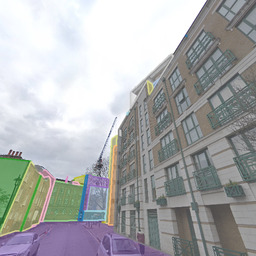}\includegraphics[width=.12\linewidth]{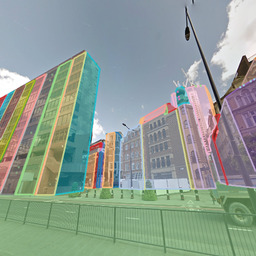}\includegraphics[width=.12\linewidth]{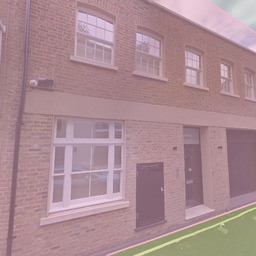}\includegraphics[width=.12\linewidth]{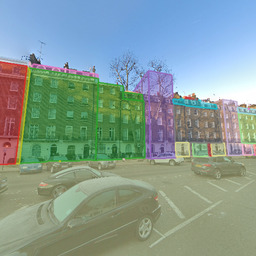}\includegraphics[width=.12\linewidth]{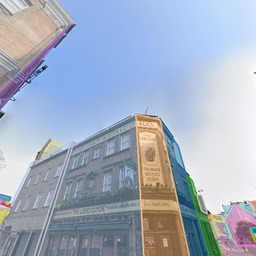}\includegraphics[width=.12\linewidth]{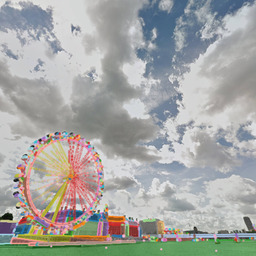}\includegraphics[width=.12\linewidth]{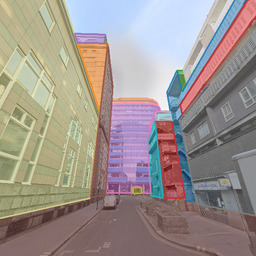}\includegraphics[width=.12\linewidth]{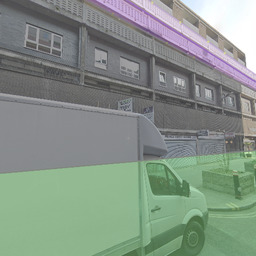}\\
\includegraphics[width=.12\linewidth]{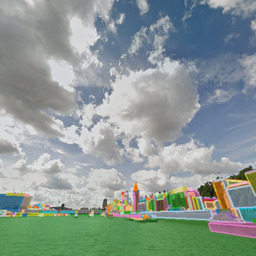}\includegraphics[width=.12\linewidth]{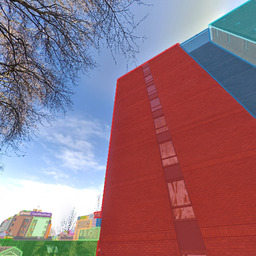}\includegraphics[width=.12\linewidth]{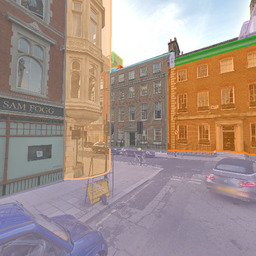}\includegraphics[width=.12\linewidth]{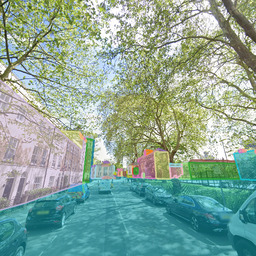}\includegraphics[width=.12\linewidth]{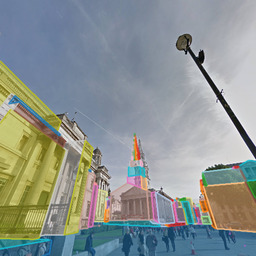}\includegraphics[width=.12\linewidth]{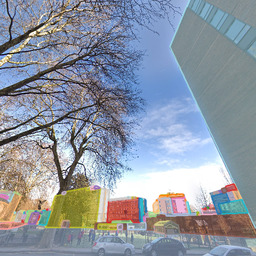}\includegraphics[width=.12\linewidth]{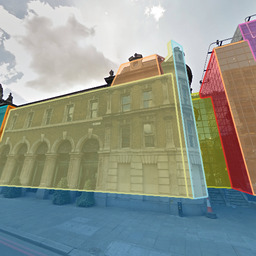}\includegraphics[width=.12\linewidth]{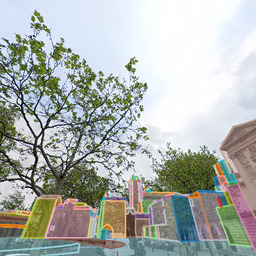}\\
\caption{\textbf{Random sampled} perspective images overlaid with surface segments from \textbf{low-resolution} CAD models.}
\label{fig:datasets:low}
\end{figure*}

\begin{figure*}[t]
\centering

\subfloat[\label{fig:annotation:pano} UI when annotating the 3D model.]{
\begin{minipage}{0.48\linewidth}
\includegraphics[width=\linewidth]{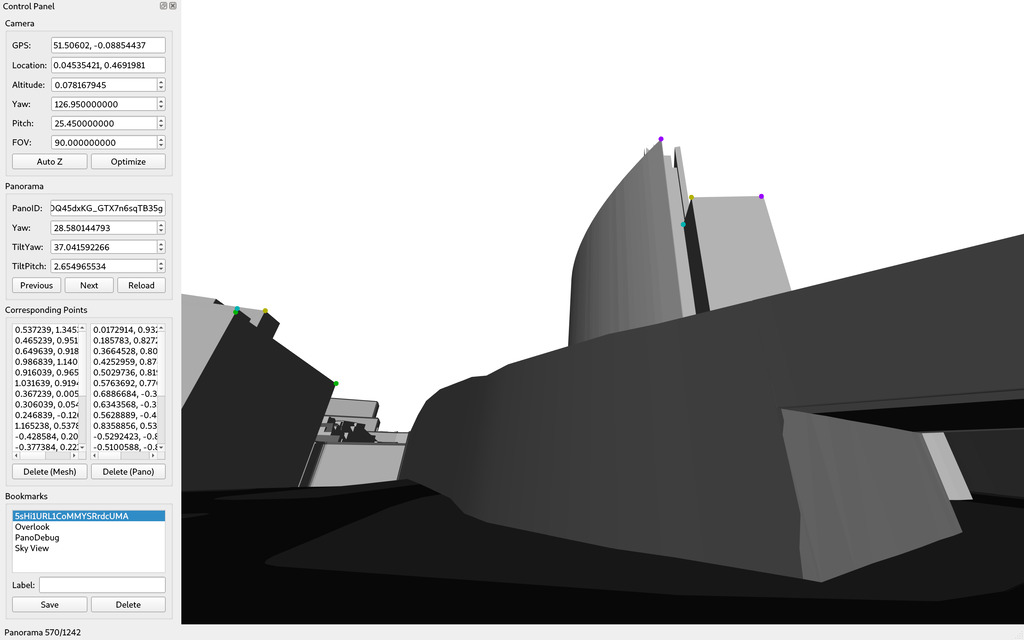}\\
\includegraphics[width=\linewidth]{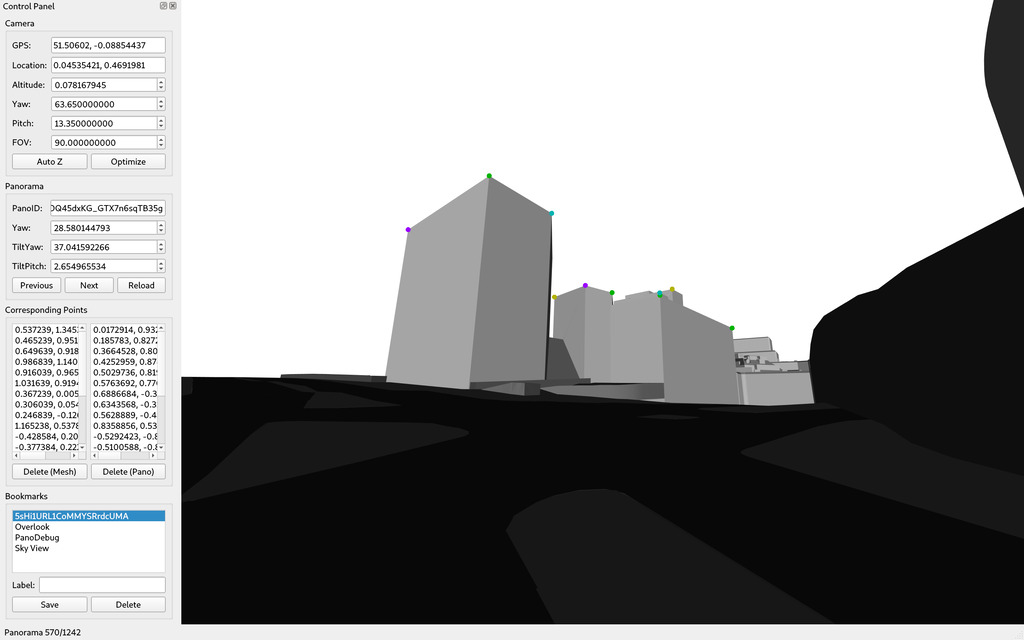}
\end{minipage}
}
\subfloat[\label{fig:annotation:Mesh} UI when annotating panorama images.]{
\begin{minipage}{0.48\linewidth}
\includegraphics[width=\linewidth]{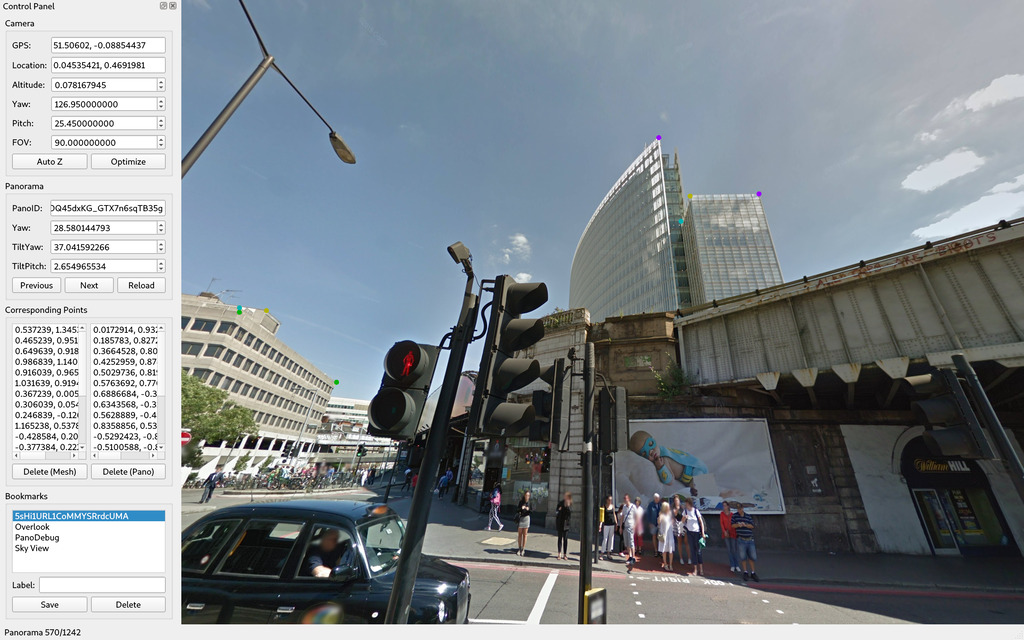}\\
\includegraphics[width=\linewidth]{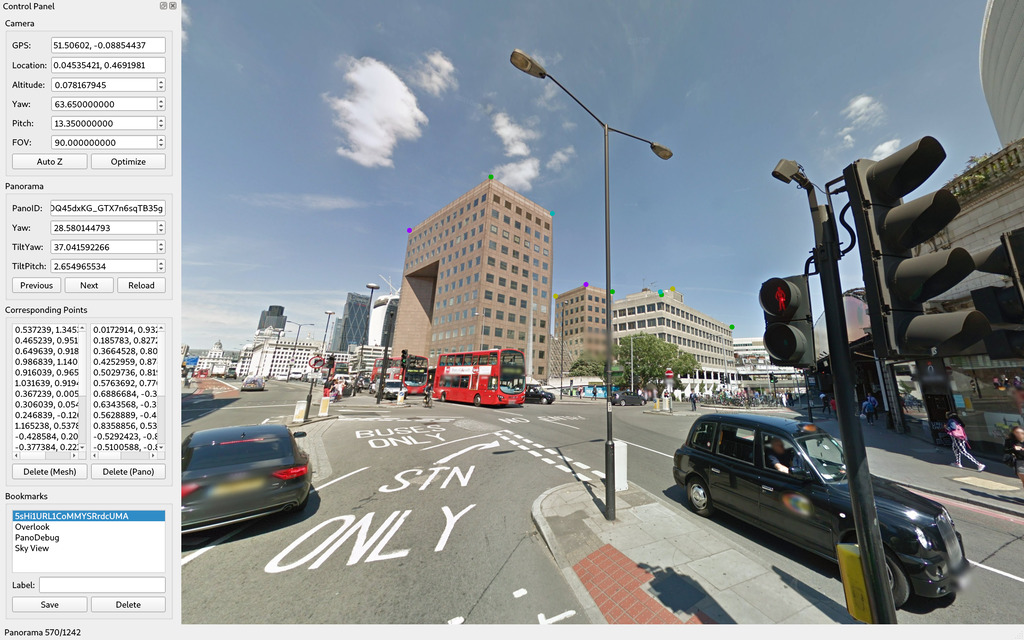}
\end{minipage}
}

\caption{User interface for panorama-to-model corresponding point annotation.\label{fig:annotation}}

\end{figure*}

In order to visualize the overall alignment quality of our dataset, we show \textbf{random sampled} images from HoliCity overlaid with the surface segmentation in \Cref{fig:datasets:high} (high-resolution CAD models) and  \Cref{fig:datasets:low} (low-resolution CAD models).
For the overlays from low-resolution CAD models, there are slightly more model errors and less details compared to the overlays from high-resolution CAD models, especially for the regions near the ground.

\subsection{Labeling Tools}

As introduced in \Cref{sec:building}, we build the correspondence between the CAD models and the panorama images through labeling pairs of corresponding points on them.
\Cref{fig:annotation} shows our labeling tool.
The annotators use our labeling tool to put points on the images and models.
Within our labeling tool, they can freely navigate in the London city with keyboard shortcuts and adjust the camera pose on the left control panel.
They can switch between 3D models and re-projected panorama images with keyboard shortcuts.
To add points to the CAD models, users could click around the vertices of the mesh and the annotation tool will automatically snap the point to the nearest visible vertices.
The users could also add points on the panorama images with mouse clicks.
When the annotator thinks he has labeled enough points, he could optimize the camera pose with current correspondence by clicking the ``Optimize'' button on the left control panel.

We instruct annotators to label points only on the corners of the building roof if possible.
This is because the CAD models are made from aerial images, and the roof features of the CAD model are usually much more reliable.
For current batches, each image contains at least 8 pairs of labeled corresponding pairs unless buildings in the panorama images are highly occluded.

\subsection{Monocular Depth Estimation}
\begin{figure}[t]
\centering
\setlength{\lineskip}{0.0ex}
\def\figsize{0.199\linewidth}

\includegraphics[width=\figsize,frame]{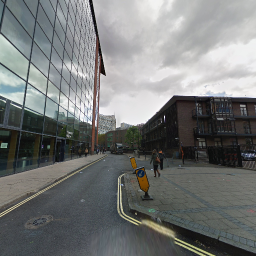}%
\includegraphics[width=\figsize,frame]{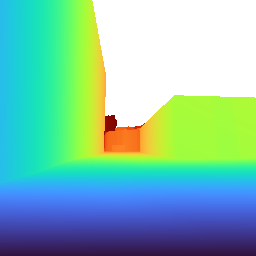}%
\includegraphics[width=\figsize,frame]{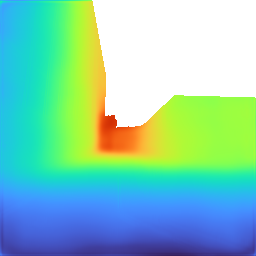}%
\includegraphics[width=\figsize,frame]{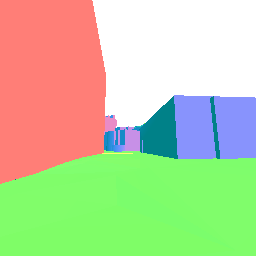}%
\includegraphics[width=\figsize,frame]{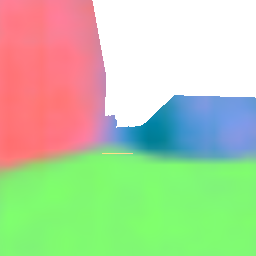}%

\includegraphics[width=\figsize,frame]{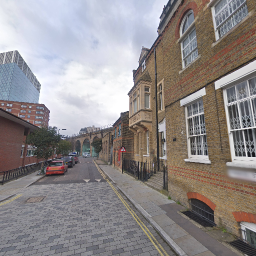}%
\includegraphics[width=\figsize,frame]{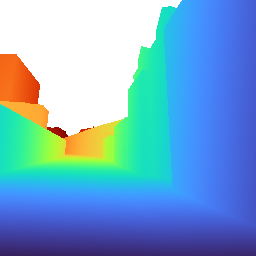}%
\includegraphics[width=\figsize,frame]{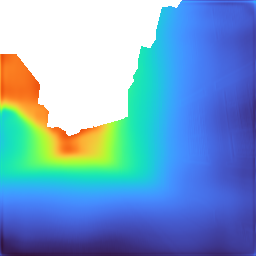}%
\includegraphics[width=\figsize,frame]{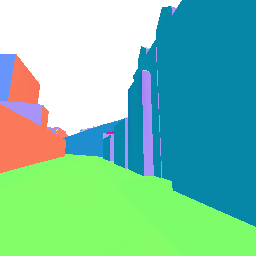}%
\includegraphics[width=\figsize,frame]{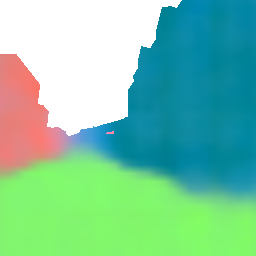}%

\vspace{2mm}
\scriptsize

\begin{minipage}{\figsize} \centering (a) RGB image \\  (input) \end{minipage}%
\begin{minipage}{\figsize} \centering (b) Ground truth \\ (depth) \end{minipage}%
\begin{minipage}{\figsize} \centering (c) Inferred \cite{ronneberger2015u}\\ (depth) \end{minipage}%
\begin{minipage}{\figsize} \centering (d) Ground truth\\ (normal) \end{minipage}%
\begin{minipage}{\figsize} \centering (e) Inferred \cite{ronneberger2015u}\\ (normal) \end{minipage}%

\caption{Visualization of results on tasks of monocular normal and depth estimation. Models are trained and evaluated on HoliCity. }
\label{fig:results:normaldepth}

\end{figure}

Monocular depth estimation has been a popular task since the beginning of deep learning \cite{eigen2015predicting}.
To demonstrate our work can better support this task than synthetic datasets, we run the following benchmarks:
We train UNet \cite{ronneberger2015u} on HoliCity and SYNTHIA (synthetic outdoor) and test them on the validation splits of HoliCity, SYNTHIA, and scene 162 of MegaDepth.
We show the qualitative and quantitative results in \Cref{fig:results:normaldepth} and \Cref{tab:results:depth}.
The goal is to compare the generalizability of HoliCity-trained models and the SYNTHIA-train models.
We report the scale-invariant error (SIL) \cite{eigen2015predicting} because depth maps of MegaDepth only have relative scales due to the usage of SfM.
Here we have two observations.
First, methods evaluated on HoliCity has larger errors than the methods evaluated on SYNTHIA.
This might be because scenes of HoliCity has more varieties than the scenes of SYNTHIA.
Second, methods trained on HoliCity has better performance when tested on other outdoor datasets (MegaDepth) than methods trained on SYNTHIA.
This shows that HoliCity-trained models have better generalizability than that of SYNTHIA-trained models, which is probably because images in HoliCity are more realistic and versatile than the images from the synthetic dataset.

\begin{table}[t]
  \resizebox{\linewidth}{!}{%
    \begin{tabular}{c|ccc}
      \hline
      \hline
      \multirow{2}{*}{Training Datasets} & \multicolumn{3}{c}{Testing Datasets (SIL \cite{eigen2015predicting} Error)} \\ \cline{2-4} 
        & HoliCity        & SYNTHIA        & MegaDepth        \\ \hline
        HoliCity                                                                     & 0.101          & 0.237       & 0.088            \\
        SYNTHIA                                                                      & 0.353          & 0.054       & 0.246            \\ \hline
      \hline
    \end{tabular}%
  }
  \caption{
    Results of monocular depth estimation cross-trained and evaluated on the validation splits of HoliCity, SYNTHIA, and scene 162 of MegaDepth.
  }
  \label{tab:results:depth}
\end{table}

\begin{figure}[tbp]
  \centering
  \hfill
  \includegraphics[width=.48\linewidth]{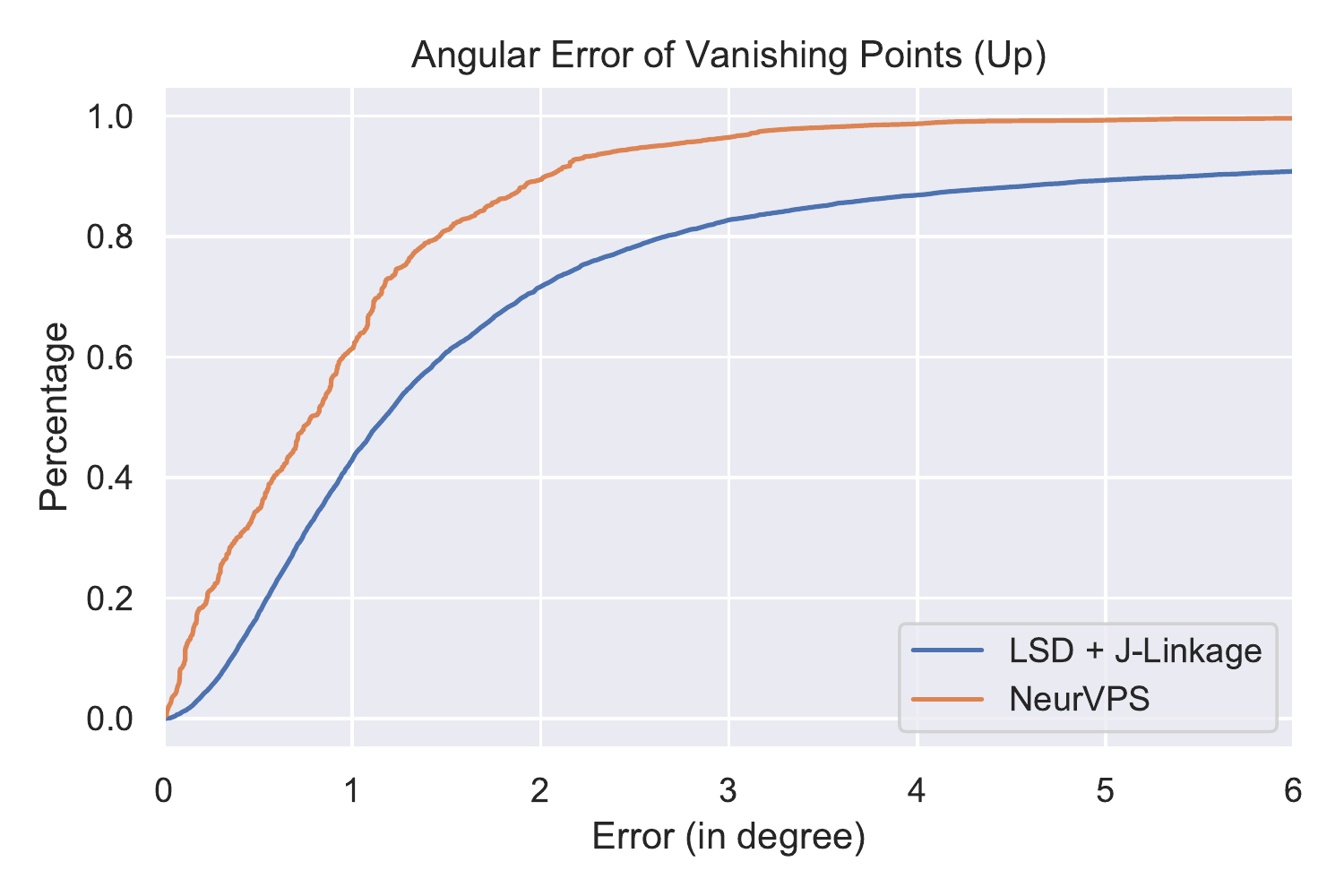}
  \hfill
  \includegraphics[width=.48\linewidth]{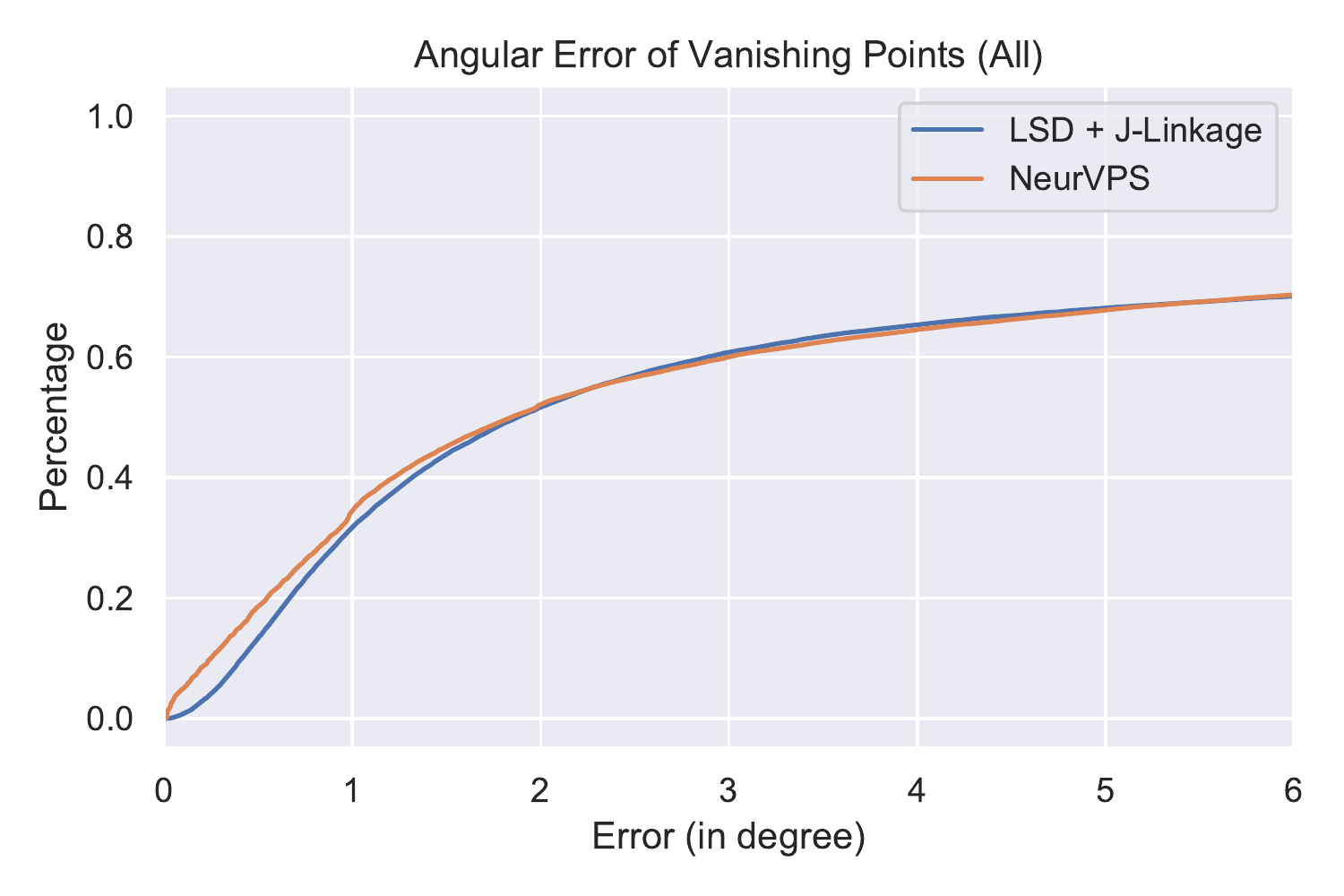}
  \hfill
  \caption{
    Angular errors of vanishing points. The left figure shows the plot of angular errors vs.
    percentages of algorithms when predicting the upward vanishing points, while the right figure
    shows the plot of angular errors vs.  percentages of algorithms when predicting all the vanishing
    points.
  }
  \label{fig:vpts}
\end{figure}

\subsection{Vanishing Point Detection} 

Vanishing points are an important concept in 3D vision, especially in outdoor urban environments.
This is because vanishing points provide information about camera poses with respect to local building structures.
Generating vanishing points with CAD models is relatively easy and accurate.
First, because all the scene contains the upward vertical vanishing points, we compute them by projecting the up vector $(0, 0, 1)$ into the image with its camera pose.
Second, we need to find horizontal vanishing points.
We solve this by clustering the surface normal with DBSCAN \citesupp{birant2007st} and find the direction of horizontal vanishing points by computing the cross product of the surface normal and the up vector.
We use DBSCAN because it does not require a predetermined number of clustering centers, and we want the number of horizontals vanishing variable.
This avoids the limitation of the strong Manhattan assumption that is used by previous datasets including YUD \citesupp{denis2008efficient}, ECD \citesupp{barinova2010geometric}, and HLW \citesupp{workman2016horizon}.

We test two algorithms, the conventional LSD \citesupp{von2010lsd,feng2010semi} + J-Linkage \cite{toldo2008robust} and the recent learning-based NeurVPS \cite{zhou2019neurvps}.
The former uses line segment detectors to detect the lines and then clusters them according to their intersection using J-Linkage.
NeurVPS, on the other hand, uses a coarse-to-fine strategy and tests whether a vanishing point candidate is valid with a conic convolutional neural network.
\Cref{fig:vpts} shows the results of both algorithms.
The median error of NeurVPS is around 0.5 degrees for the up vanishing points and 1.5 degrees for all vanishing points.
This shows that the camera pose of our dataset should at least be around that accuracy.
Besides, we find that learning-based NeurVPS has the similar performance as LSD when considering all the vanishing points.
This is probably because NeurVPS has not yet been optimized for detecting a variable number of vanishing points.

\subsection{Relocalization}
Precise camera localization from images is key to many 3D vision tasks, such as visual compasses, navigation, autonomous driving, and augmented reality.
Due to the cost, existing outdoor datasets only capture limited regions~\citesupp{shotton2013scene} or a few paths~\cite{maddern20171}.
In comparison, \OURS{} provides an accurate camera pose densely sampled in an entire city with various view angles.
We believe that a method with decent performance on our dataset will be useful for many real-world applications.
Here, we benchmark state-of-the-art methods on our dataset and find a significant gap between the existing techniques and real-world challenges.

\begin{figure}[tb]
    \centering
    \hfill
    \includegraphics[width=.48\linewidth]{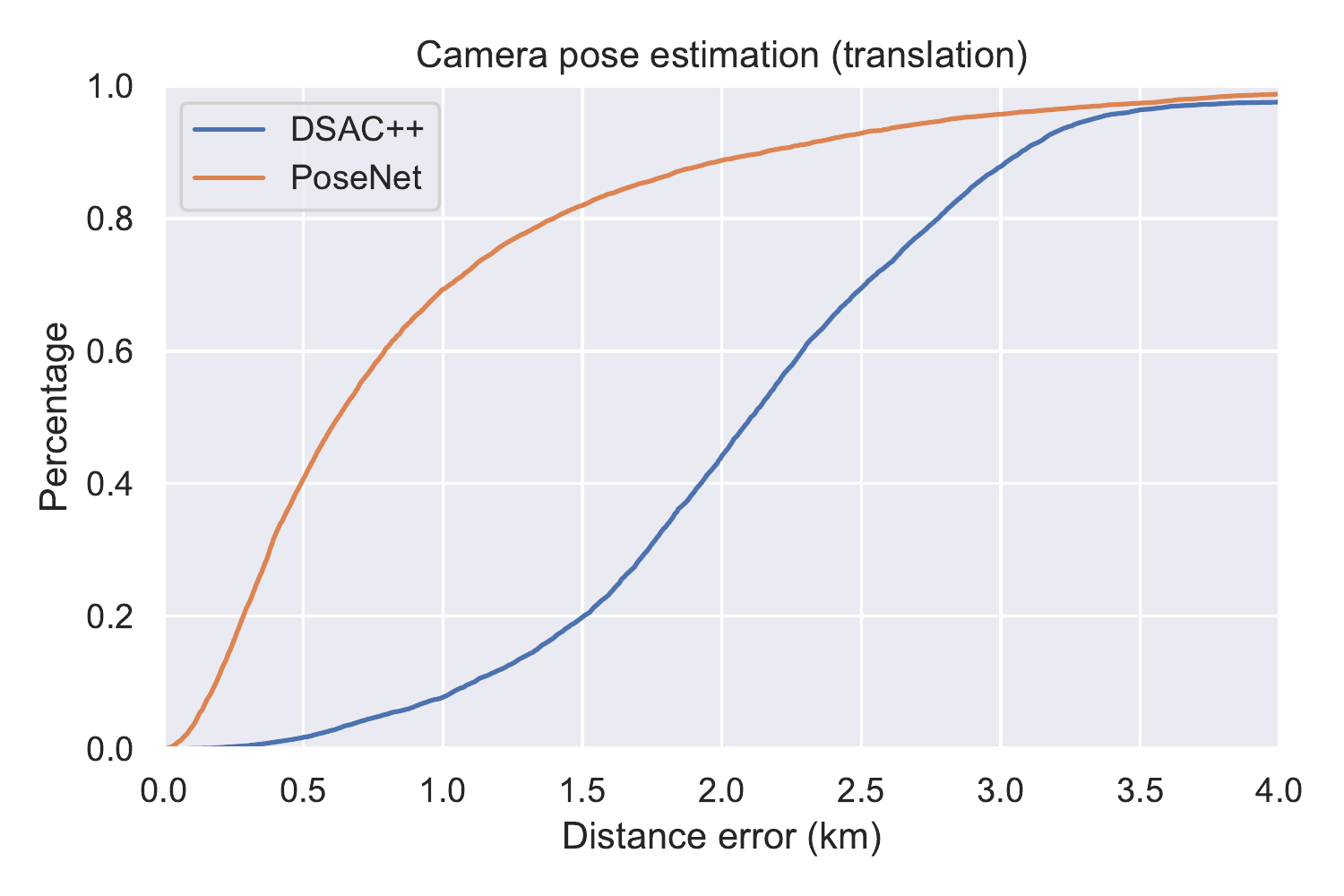}
    \hfill
    \includegraphics[width=.48\linewidth]{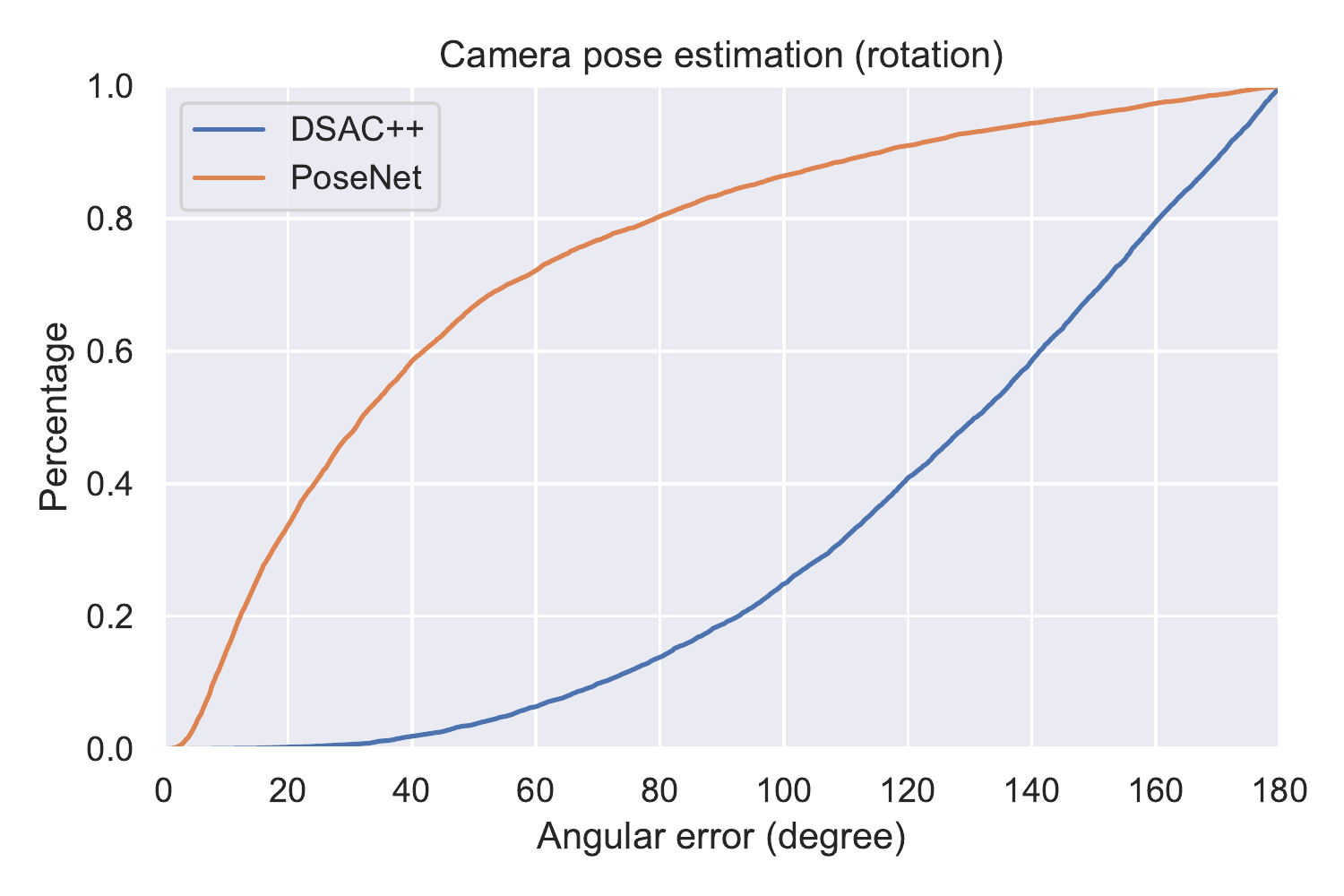}
    \hfill
    \caption{Orientation and Localization errors for existing camera localization methods trained and tested using \OURS{} dataset.}
    \label{fig:localization-statistics}
\end{figure}

We choose to evaluate two recent deep learning-based methods, namely PoseNet \citesupp{kendall2017geometric} and DSAC++ \citesupp{brachmann2018learning} as two representative works related to direct regression and scene coordinate estimation.
\Cref{fig:localization-statistics} summarizes the errors in the camera distance and orientation predicted.
The Y-axis represents the threshold of the prediction error and the X-axis is the percentage of frames in the test data with prediction error smaller than the threshold.
The mean errors in location for PoseNet and DSAC++ are $921$m and $2,086$m, respectively. The mean orientation errors for them are $47.1^{\circ}$ and $124.2^{\circ}$, respectively.
Although such results look problematic, we do try our best to tune the parameters of the networks and find it hard to reach a reasonable performance on HoliCity.
In addition, we do observe that in the original paper \citesupp{kendall2017geometric}, their results on scenes such as ``Office'' have similar accuracy as ours.
We think that the difficulty of HoliCity for these deep learning-based relocalization comes from its massive scale (20km${}^2$), relatively large baseline (\Cref{fig:teaser:satellite}), and long span (\Cref{sec:exploring}), hence resulting in huge prediction errors for these learning-based relocalization approaches.

{
  \bibliographystylesupp{ieee_fullname}
  \bibliographysupp{main}
}

\end{document}